\documentclass[letterpaper]{article} 
\usepackage{aaai2026}  
\usepackage{times}  
\usepackage{helvet}  
\usepackage{courier}  
\usepackage[hyphens]{url}  
\usepackage{graphicx} 
\usepackage{natbib}  
\usepackage{caption} 
\usepackage{algorithm}
\usepackage{algorithmic}

\usepackage{newfloat}
\usepackage{listings}
\DeclareCaptionStyle{ruled}{labelfont=normalfont,labelsep=colon,strut=off} 
\floatstyle{ruled}
\newfloat{listing}{tb}{lst}{}
\floatname{listing}{Listing}
\nocopyright

\title{Improving Densification in 3D Gaussian Splatting for High-Fidelity Rendering}
\author{
    Xiaobin Deng,
    Changyu Diao\corresponding,
    Min Li,
    Ruohan Yu,
    Duanqing Xu\corresponding
}
\affiliations{
    Zhejiang University
}

\usepackage{bibentry}

\begin{document}

\maketitle

\begin{abstract}
    Although 3D Gaussian Splatting (3DGS) has achieved impressive performance in real-time rendering, its densification strategy often results in suboptimal reconstruction quality.
    In this work, we present a comprehensive improvement to the densification pipeline of 3DGS from three perspectives: when to densify, how to densify, and how to mitigate overfitting.
    Specifically, we propose an Edge-Aware Score to effectively select candidate Gaussians for splitting.
    We further introduce a Long-Axis Split strategy that reduces geometric distortions introduced by clone and split operations.
    To address overfitting, we design a set of techniques, including Recovery-Aware Pruning, Multi-step Update, and Growth Control.
    Our method enhances rendering fidelity without introducing additional training or inference overhead, achieving state-of-the-art performance with fewer Gaussians.
\end{abstract}


Code: https://xiaobin2001.github.io/improved-gs-web
\section{Introduction}
Novel view synthesis (NVS) is a classical problem in computer vision, with widespread applications in virtual reality, cultural heritage preservation, autonomous driving, and other fields.
Neural Radiance Field (NeRF) \cite{mildenhall2021nerf} introduced the use of neural networks to learn the structure and features of a scene, requiring only multi-view 2D images as training data to synthesize novel views.
However, NeRF suffers from long synthesis times for individual views \cite{muller2022instant, fridovich2022plenoxels}, making real-time rendering challenging.
Recently, 3D Gaussian Splatting (3DGS) \cite{kerbl20233d} has attracted attention due to its explicit representation and real-time rendering performance.

3DGS represents a scene using a large number of 3D Gaussian ellipsoids.
The properties of these Gaussians include position, shape, opacity, and color, all of which can be optimized through differentiable rendering.
3DGS generates an initial set of Gaussians from the sparse points obtained through Structure from Motion (SfM) \cite{schonberger2016structure}, and subsequently refines the scene representation by adding Gaussians via adaptive density control (ADC).

The densification of 3D Gaussian Splatting has a significant impact on rendering quality, However, the ADC yields unsatisfactory performance.
To achieve high-fidelity rendering, we conduct a comprehensive analysis and improvement of the densification strategy in 3DGS, addressing three key aspects: when to densify, how to densify, and how to mitigate overfitting.
\textbf{When to densify:} In 3DGS, view-averaged coordinate gradients are used to select Gaussians for densification.
However, due to gradient conflicts\cite{ye2024absgs}, this strategy fails to identify certain large Gaussians that contribute to blurred reconstructions.
To select higher-quality candidates, we introduce an Edge-Aware Score (EAS) approach, which combines edge information with pixel-level loss.
\textbf{How to densify: } In 3DGS, the shape of Gaussians gradually adapts to the scene through backpropagation-based optimization.
However, the clone and split operations induce sudden geometric perturbations in the regions previously represented by the original Gaussians.
Minimizing such geometric discrepancies can clearly improve optimization efficiency.
Inspired by this insight, we propose Long-Axis Split(LAS), which carefully designs the relative positions, shapes, and opacities of child Gaussians to minimize the geometric difference before and after splitting.
\textbf{Mitigating overfitting:} We introduce three techniques to alleviate overfitting:
1. Recovery-Aware Pruning(RAP): This method removes potentially overfitted Gaussians early during the densification phase.
2. Multi-step Update(MU): During training after the densification process, we perform parameter updates every $N(N>1)$ iterations.
3. Growth Control(GC): A smooth curve is introduced to constrain the growth rate of Gaussian counts.

In summary, our method significantly improves the rendering quality of 3DGS without introducing additional training or rendering overhead.
Compared to 3DGS and the current state-of-the-art works, our approach achieves better rendering quality while utilizing fewer Gaussians.
Our contributions are summarized as follows:
\begin{itemize}
  
      \item We propose an Edge-Aware Score approach to better identify candidates for densification.
      \item We propose Long-Axis Split, which is designed to minimize geometric inconsistencies introduced during the densification process. 
      \item We present multiple techniques to mitigate overfitting.
      \item Our work significantly improves the rendering quality of 3DGS without introducing additional computational overhead.
      
\end{itemize}

\section{Related Works}
3D Gaussian Splatting (3DGS) \cite{kerbl20233d} demonstrates outstanding performance in both rendering quality and speed, representing the current state-of-the-art in novel view synthesis.
3DGS has been widely adopted across fields including dynamic scenes \cite{wu20244d, lin2024gaussian}, simultaneous localization and mapping (SLAM) \cite{matsuki2024gaussian}, 3D content generation \cite{chen2024text}, autonomous driving \cite{zhou2024drivinggaussian}, and high-fidelity human avatars \cite{shao2024splattingavatar}.

Numerous studies have focused on improving 3DGS rendering quality.
For instance, Mip-Splatting \cite{yu2024mip} introduces a 3D smoothing filter and a 2D mipmap filter to eliminate aliasing artifacts present in 3DGS during scaling.
To mitigate the impact of defocus blur on reconstruction quality, Deblurring 3DGS \cite{lee2024deblurring} applies a small multi-layer perceptron (MLP) to the covariance matrix, learning spatially varying blur effects.
GaussianPro \cite{cheng2024gaussianpro} leverages optimized depth and normal maps to guide densification, filling gaps in areas initialized via SfM.
Spec-Gaussian \cite{yang2024spec} employs anisotropic spherical Gaussian appearance fields for Gaussian color modeling, enhancing 3DGS rendering quality in complex scenes with specular and anisotropic surfaces.
Notably, all these enhancements rely on the original density control and could benefit from our proposed work.

There are many works dedicated to improving the densification strategies of 3DGS.
Mini-Splatting \cite{fang2024mini} addresses this by generating depth maps for trained scenes to reinitialize the sparse points, and identifies blurred Gaussians with large rendering areas during training, splitting them as needed.
Pixel-GS \cite{zhang2024pixel} reduces blur through pixel-area-weighted averaging of gradients across views.
AbsGS \cite{ye2024absgs} attribute blur in reconstructions to conflicts in gradient direction across pixels when computing Gaussian coordinate gradients.
This conflict leads to larger Gaussians, which represent blur, receiving insufficient average gradients.
To resolve this, they compute Gaussian coordinate gradients by taking the modulus of pixel coordinate gradients before summing.
TamingGS \cite{mallick2024taming} proposes a densification judgment condition that employs a weighted combination of multiple scores.
RevisingGS \cite{rota2024revising} optimizes the opacity bias of Gaussians after cloning, and also uses pixel loss as a criterion for selecting candidate points. 
3DGS-MCMC \cite{kheradmand20243d} treats the insertion and optimization of Gaussians as a Stochastic Gradient Langevin Dynamics procedure.
SteepGS \cite{wang2025steepest}  designs an algorithm to compute a splitting matrix that determines whether a Gaussian should be split and where the child Gaussians should be placed.
Perceptual-GS \cite{zhou2025perceptual} introduces multi-view perceptual sensitivity to guide densification as well as parameter optimization.

We compare against all the above-mentioned works except RevisingGS, which has not made its code publicly accessible.
Experiments demonstrate that our method surpasses all the aforementioned works in rendering quality.
\section{Methods}

\subsection{Preliminaries}

3DGS defines the scene as a set of anisotropic 3D Gaussian primitives:
\begin{equation}
    G(x)=\exp{\left( -\frac{1}{2} (x)^T \Sigma^{-1} (x) \right)},
\end{equation}
where $\Sigma$ is the 3D covariance matrix and $x$ represents the position relative to the Gaussian mean coordinates.
To ensure the semi-definiteness of the covariance matrix, 3DGS reparameterizes it as a combination of a rotation matrix $R$ and a scaling matrix $S$:
\begin{equation}
    \Sigma = R S S^T R^T.
\end{equation}
The scaling matrix $S$ can be represented using a 3D vector $s$, while the rotation matrix $R$ is obtained from the quaternion $q$.
To render an image from a specified viewpoint, the color of each pixel $p$ is obtained by blending $N$ ordered Gaussians $ \{ G_i \mid i = 1, \dots, N \}$ that cover pixel $p$, with the following formula:
\begin{equation}
	\label{eq:front-to-back}
	C = \sum_{i=1}^{N}
	c_{i}\alpha_{i}
	\prod_{j=1}^{i-1}(1-\alpha_{j}),
\end{equation}
where $\alpha_{i}$ is the value obtained by projecting $G_i$ onto $p$ and multiplying by the opacity of $G_i$, while $c_{i}$ represents the color of $G_i$, expressed by SH coefficients.

3DGS initializes the scene using sparse points generated by SfM, and then increases the number and density of Gaussians in the scene through adaptive density control.
Specifically, 3DGS calculates the cumulative average view-space positional gradients of Gaussians every 100 iterations, with each iteration training a single viewpoint.
The formula for calculating the average gradient is as follows:
\begin{equation}
    \frac{\sum_{k=1}^{M^i}{\sqrt{\left( \frac{\partial L_k}{\partial \mu _\mathrm{k,x}^{i}} \right) ^2+\left( \frac{\partial L_k}{\partial \mu _\mathrm{k,y}^{i}} \right) ^2}}}{M^i}>\tau _\mathrm{pos},
\end{equation}
where $M^i$ represents the number of viewpoints in which the Gaussian participates during a cycle, $\tau _\mathrm{pos}$ is the given densification threshold, 
$\frac{\partial L_k}{\partial \mu _\mathrm{k,x}^{i}}$ and $\frac{\partial L_k}{\partial \mu _\mathrm{k,y}^{i}}$ represent the gradients of the Gaussian with respect to the $x$ and $y$ for the current viewpoint, obtained by summing the gradients of each pixel with respect to the coordinates:
\begin{equation}
    \frac{\partial L_k}{\partial \mu _\mathrm{k,x}^{i}} = \sum_{j=1}^m  \frac{\partial L_j}{\partial \mu_{i,x}}, \ 
    \frac{\partial L_k}{\partial \mu _\mathrm{k,y}^{i}} = \sum_{j=1}^m  \frac{\partial L_j}{\partial \mu_{i,y}}.
\end{equation}
Gaussians with average gradients exceeding a predefined threshold undergo densification using either clone or split, depending on their size.

The ADC strategy mainly includes three operations.
\textbf{Clone}: Duplicate a small Gaussian with parameters (including position) identical to the parent.
Since the clone operation occurs after the rendering step, the cloned Gaussian does not receive gradients during the current iteration.
In subsequent parameter updates, only the parent Gaussian’s parameters are modified.
\textbf{Split}: A large Gaussian is replaced by two smaller Gaussians, which retain the same shape, opacity, and color as the original.
Each smaller Gaussian is scaled down to $1/1.6$ of the parent's size.
The coordinates of the two smaller Gaussians are generated through Gaussian sampling, using the parent’s position and covariance matrix as parameters.
\textbf{Reset Opacity}: During densification, an opacity reset operation is performed every $3000$ iterations.
Specifically, the opacity of Gaussians with opacity greater than $0.01$ is reset to $0.01$.

\subsection{Edge-Aware Score}

The densification process aims to enhance reconstruction quality by increasing the density of Gaussians.
Human vision is particularly sensitive to blurring in edge regions, yet conventional pixel-wise losses fail to adequately capture these areas.
To address this limitation, we propose an edge-aware scoring mechanism.

First, we apply the Laplacian operator to each training image for edge detection, generating per-pixel edge weights.
Then, for each Gaussian, we compute its edge-aware score under a given view using the following formula:
\begin{equation}
S_{i,j} = \sum_{p \in \mathcal{P}} {\omega _\mathrm{p,j}^{i}}  \cdot {\alpha _\mathrm{p,j}^{i}}
\end{equation}
where $ \omega _\mathrm{p,j}^{i} $ denotes the edge weight of pixel $ p $ in view $ j $, and $ \alpha _\mathrm{p,j}^{i} $ represents the rendering weight of Gaussian $ i $ at pixel $ p $ in view $ j $.
Before each densification step, we randomly sample $ N_s $ views from the training set and compute the average edge-aware score $ \bar{S}_i $ of each Gaussian across these views:
\begin{equation}
\bar{S}_i = \frac{1}{N_s} \sum_{j=1}^{N} S_{i,j}
\end{equation}

The edge-aware score represents the contribution of a Gaussian to image edge regions, but it is not sufficient to identify the rendering quality of the region in which it resides.
To address this issue, we use the view-averaged absolute coordinate gradients as the second criterion for evaluation. The absolute coordinate gradient is obtained by replacing Equation~(4) with:
\begin{equation}
\frac{\partial L_k}{\partial \mu _\mathrm{k,x}^{i}} = \sum_{j=1}^m  
\left| \frac{\partial L_j}{\partial \mu_{i,x}} \right|,
\quad
\frac{\partial L_k}{\partial \mu _\mathrm{k,y}^{i}} = \sum_{j=1}^m  
\left| \frac{\partial L_j}{\partial \mu_{i,y}} \right|.
\end{equation}
The absolute gradient effectively approximates the pixel-wise loss and can be used to evaluate the rendering quality of the region occupied by the Gaussian.
Gaussians with an average absolute gradient greater than a specified threshold are selected as candidates.
The probability of splitting a candidate Gaussian is proportional to its average edge-aware score.

\subsection{Long-Axis Split}

Minimizing the geometric disturbance introduced during the densification process can accelerate optimization speed and sometimes even improve the final rendering quality.
The split operation in 3DGS generates the relative positions of child Gaussians using probabilistic sampling, which introduces randomness and limits fine-grained shape control.
In our approach, we fix the positions of the child Gaussians to be symmetric along the longest axis of the original Gaussian, with the original center as the midpoint.
Under this configuration, the position of each child Gaussian is solely determined by the distance from its center to the original Gaussian center.
The three axes of the child Gaussians are initialized with the same values and then gradually optimized to become anisotropic.
When a rendering ray is parallel to one of the axes, that axis will not contribute to the rendering, so the longest axis has the minimal probability of being aligned with the rendering rays during projection.
Therefore, we choose the longest axis as the splitting direction to maximize the rendering contribution.

\begin{figure}[ht]
    \centering
    \includegraphics[width=0.47\textwidth]{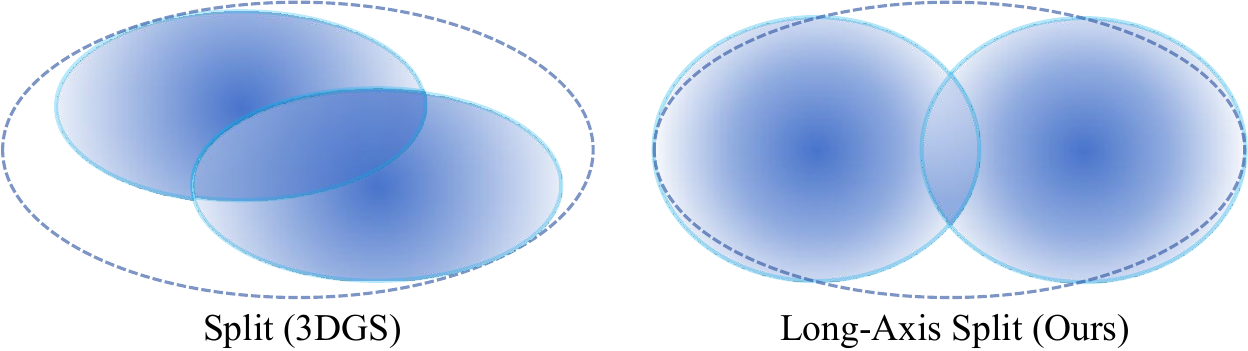}
    \caption{Compare the geometric differences before and after splitting between split and LAS. The outer dashed line represents the Gaussian shape before splitting.}
    \label{fig:LAS-shape}
\end{figure}

Let $d$ denote the distance from the center of a child Gaussian to the original Gaussian center, and let $L_0$ be the semi-length of the original Gaussian's longest axis.
The semi-length $L_s$ of the child’s longest axis should be adjusted to $L_0 - d$, ensuring that the two child Gaussians are tangent to the original Gaussian at its endpoints.
As $d$ decreases from $L_0$ to $0.5L_0$, the shape difference between before and after the split persists.
Further reducing $d$ from $0.5L_0$ to $0$ reduces the shape difference even more but increases the overlapping area between the two child Gaussians.

To determine $d$ (with the constraint $ d \leq 0.5L_0 $) while minimizing the shape difference, the lengths of the other two axes of the child Gaussians, denoted $ R_s $, should satisfy:

\begin{equation}
R_s = R_0 \cdot \sqrt{1 - \frac{d^2}{L_0^2}}
\end{equation}

This ensures that the endpoints of the two minor axes of the child Gaussians lie exactly on the surface of the original Gaussian.
A detailed proof is provided in the Appendix A.4.
Gaussian overlap may interfere with their individual optimization.
However, the rendering weights near the edges of Gaussians are significantly lower than those at the centers, making small overlaps at the edges acceptable.
In practice, we set $ d = 0.45L_0 $ to balance the trade-off between minimizing overlap and maintaining low shape difference (see Figure~\ref{fig:LAS-shape}).
See Appendix A.8 for the ablation study.

\begin{figure}[ht]
    \centering
    \includegraphics[width=0.45\textwidth]{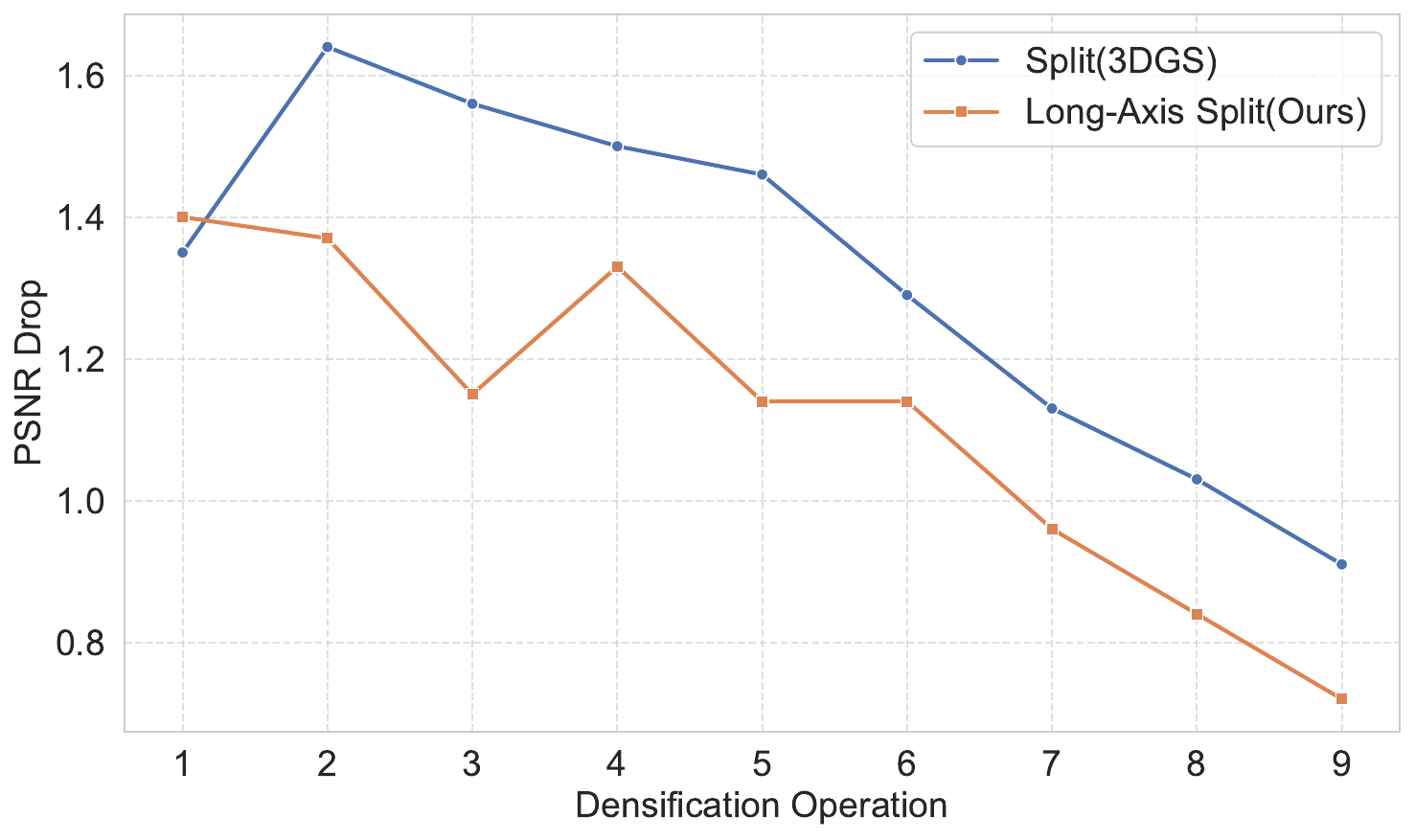}
    \caption{Evaluate the drop in PSNR after splitting using different splitting strategies, test scene is bicycle, a smaller drop indicates less geometric error introduced by the splitting.}
    \label{fig:PSNR_Drop}
\end{figure}

After split, the coverage area of the original Gaussian transitions from a single-center distribution to a dual-center distribution, which introduces a discrepancy from the perspective of density distribution.
Reducing the opacity of the child Gaussians appropriately can help mitigate this discrepancy.
In practice, we set the opacity of the child Gaussians to 60\% of the original opacity. 3DGS uses clone and split to address under-reconstruction and over-reconstruction.
However, during optimization, the size of Gaussians tends to converge to a value that balances under-reconstruction and over-reconstruction to minimize loss.
Therefore, we only use Long-Axis Split as the densification operation.
Figure~\ref{fig:PSNR_Drop} shows that compared to split, the splitting error introduced by LAS is smaller.

\subsection{Recovery-Aware Pruning}

During training, 3DGS may generate some overfitted Gaussians that contribute less from the training viewpoints but negatively impact generalization performance.
The original 3DGS resets the opacity at iterations 3K, 6K, 9K, and 12K.
By leveraging the difference in opacity recovery speed, we can eliminate some of these overfitted Gaussians.
Specifically, we prune the bottom 20\% Gaussians in terms of opacity at iteration 3300 and 6300.
Since the densification process continues until iteration 15K and RAP is only applied in the early stages, it does not affect the final number of Gaussians or the training speed.

\subsection{Muti-view Update}
In 3DGS, each training iteration processes and updates parameters based on a single view, i.e., the batch size is 1.
Increasing the batch size can improve generalization performance; however, training on multiple views per iteration significantly increases the training cost.
In contrast, indirectly increasing the effective batch size by extending the parameter update interval ($N>1$) can reduce training overhead.
Unfortunately, this approach leads to a notable degradation in final rendering quality (see Appendix A.6). This is due to two main reasons:
\begin{itemize}
      \item The densification phase requires rapid optimization after splitting. Otherwise, Gaussians that have not been sufficiently optimized may be incorrectly split again.
      \item Extending the update interval reduces the frequency of gradient updates, thereby slowing down the convergence of the parameters.
\end{itemize}
We observe that the optimization speed naturally slows down after the densification phase, which reduces the dependency on frequent updates.
At this stage, extending the update interval no longer causes erroneous splits. Therefore, we propose a two-stage training strategy:
\begin{itemize}
      \item In the early training phase, we maintain single-view updates to meet the optimization requirements during densification.
      \item After densification is complete (specifically, between 15,000 and 22,500 iterations, we use N=5, followed by N=20 afterward), we switch to multi-view batching. This enables joint optimization of computational efficiency and generalization performance.
\end{itemize}

\subsection{Growth Control}

When using the Edge Awareness Score to select split candidates, we observed that the number of Gaussians reaches its peak early in the densification stage.
An early peak in Gaussian count increases the risk of overfitting and prolongs the overall training time.
To address this, we employ a smooth convex curve as the control curve for the number of Gaussians, ensuring it meets two requirements: rapid densification in the early stages to maintain training efficiency, and a peak in Gaussian count only near the end of the densification phase.
Specifically, the Gaussian budget $N$ for the current training iteration is calculated using the following formula:
\begin{equation}
N = N_{max} \cdot \sqrt{ \frac{I - I_{start}}{I_{end} - I_{start}} }
\end{equation}
where  $I$ is the current round, with $I_{start}$ and $I_{end}$ representing the start and end iteration of the calculation, respectively, and $N_{max}$ is the user-defined final budget. 
Appendix A.9 shows the shape of the growth curves.
\section{Experiments}
\setlength{\tabcolsep}{2pt}
\begin{figure*}[t]
    \centering
    \includegraphics[width=0.93\textwidth]{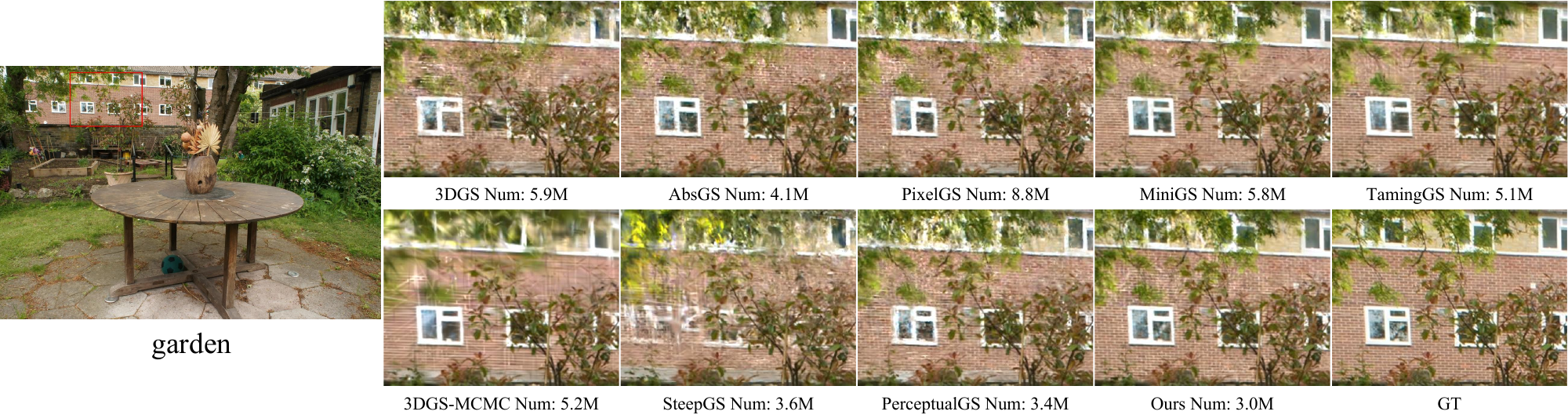} \\
    \includegraphics[width=0.93\textwidth]{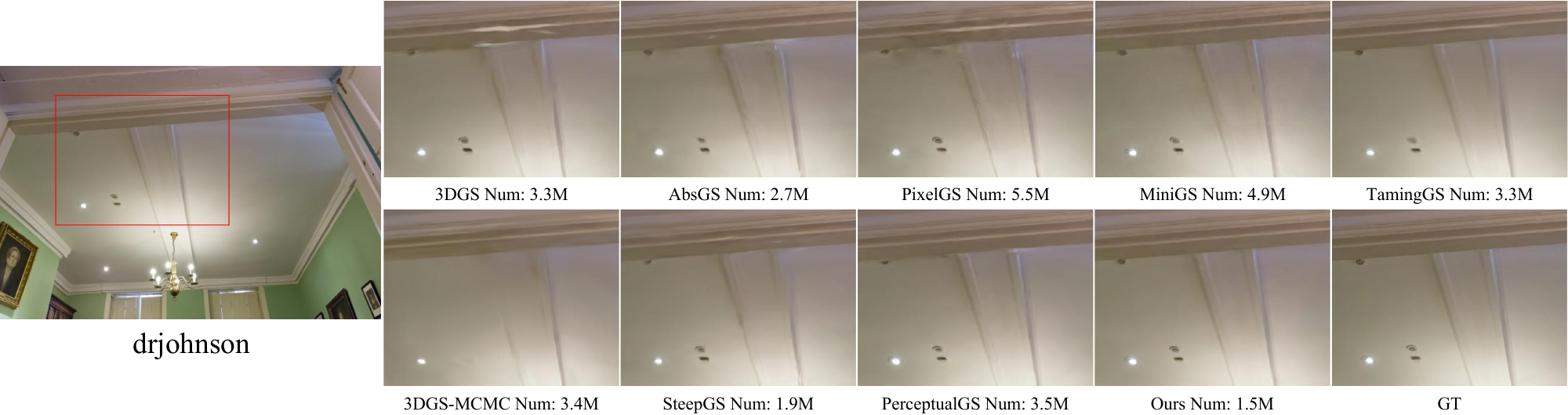} \\
    \includegraphics[width=0.93\textwidth]{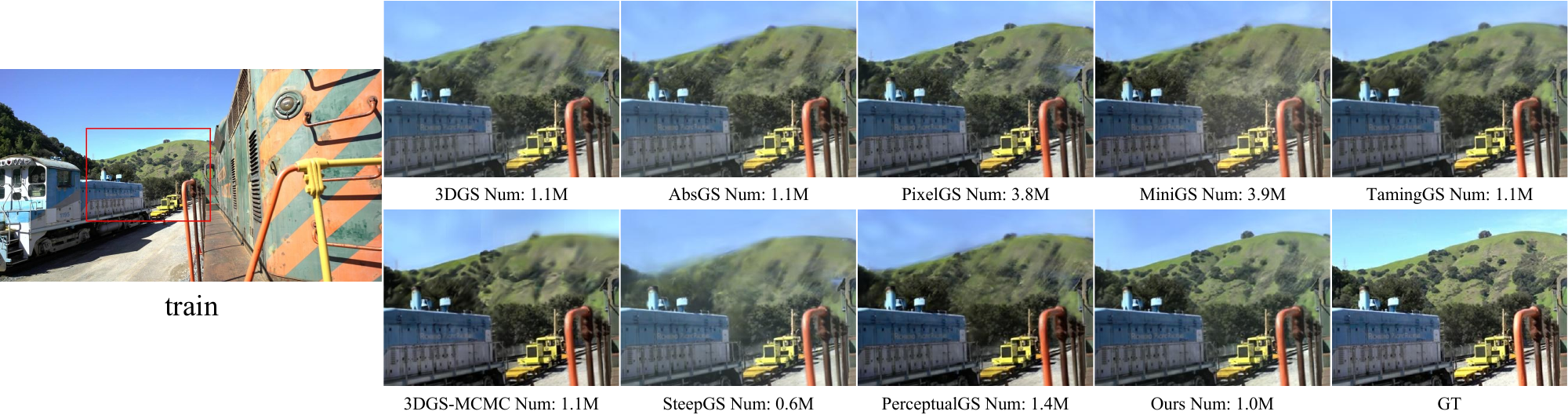}
    \caption{Qualitative comparison results among scenes garden, drjohnson, train.}
    \label{fig:qualitative_comparison}
\end{figure*}

\begin{table*}[ht]
	\centering
	\scalebox{0.79}{
		\begin{tabular}{l|cccc|cccc|cccc}
			
			Dataset & \multicolumn{4}{c|}{Mip-NeRF360}  & \multicolumn{4}{c|}{Deep Blending} & \multicolumn{4}{c}{Tanks\&Temples} \\
			Method|Metric
			& $SSIM^\uparrow$   & $PSNR^\uparrow$    & $LPIPS^\downarrow$  & $Num^\downarrow$
			& $SSIM^\uparrow$   & $PSNR^\uparrow$    & $LPIPS^\downarrow$  & $Num^\downarrow$
			& $SSIM^\uparrow$   & $PSNR^\uparrow$    & $LPIPS^\downarrow$  & $Num^\downarrow$ \\
			\hline
			3DGS& 0.815 & 27.48 & 0.216 & 3337659 & 0.904 & 29.57 & 0.244 & 2832494 & 0.848 & 23.69 & 0.177 & 1847041 \\
            \hline
            AbsGS (ACMMM24) & 0.820 & 27.52 & 0.198 & 3194225 & 0.905 & 29.49 & 0.243 & 2054043 & 0.857 & 23.83 & 0.164 & 1442984 \\
            PixelGS (ECCV24) & 0.824 & 27.62 & 0.189 & 5619828 & 0.897 & 28.98 & 0.248 & 4644260 & 0.857 & 23.84 & 0.149 & 4519926 \\
            MiniSplatting-D (ECCV24) & 0.832 & 27.57 & \colorbox{red!40}{0.176} & 4685127 & 0.906 & \colorbox{orange!40}{29.93} & \colorbox{red!40}{0.211} & 4627579 & 0.855 & 23.36 & \colorbox{red!40}{0.140} & 4260423 \\
            TamingGS (SIGGRAPHAsia24) & 0.822 & 27.96 & 0.207 & 3182444 & 0.907 & \colorbox{orange!40}{29.93} & 0.236 & 2799868 & 0.860 & \colorbox{orange!40}{24.42} & 0.163 & 1849918 \\
            3DGS-MCMC (NeurIPS24) & \colorbox{orange!40}{0.835} & \colorbox{orange!40}{28.01} & \colorbox{orange!40}{0.186} & 3227778 & \colorbox{orange!40}{0.912} & 29.78 & 0.237 & 2950000 & \colorbox{orange!40}{0.869} & 24.40 & 0.149 & 1850000 \\
            SteepGS (CVPR25) & 0.795 & 27.05 & 0.247 & \colorbox{orange!40}{2193808} & 0.905 & 29.74 & 0.251 & \colorbox{orange!40}{1605267} & 0.838 & 23.42 & 0.193 & \colorbox{orange!40}{1310323} \\
            Perceptual-GS (ICML25) & 0.829 & 27.77 & 0.187 & 2685908 & 0.906 & 29.88 & 0.231 & 2892183 & 0.857 & 23.88 & 0.150 & 1721090 \\
            \hline
            Ours & \colorbox{red!40}{0.836} & \colorbox{red!40}{28.19} & \colorbox{orange!40}{0.186} & \colorbox{red!40}{1777778} 
            & \colorbox{red!40}{0.913} & \colorbox{red!40}{30.19} & \colorbox{orange!40}{0.226} & \colorbox{red!40}{1250000} 
            & \colorbox{red!40}{0.872} & \colorbox{red!40}{24.59} & \colorbox{orange!40}{0.145} & \colorbox{red!40}{1250000}
        \end{tabular}
	}
	\caption{Quantitative results on the Mip-NeRF 360, Deep Blending, and Tanks and Temples datasets. Cells are highlighted as follows: \colorbox{red!40}{best}, and \colorbox{orange!40}{second best}. All parameters for the compared methods follow the default settings from their respective original papers.}
	\label{tab:quantitative_results}
\end{table*}

\begin{figure*}[t]
    \centering
    \includegraphics[width=0.95\textwidth]{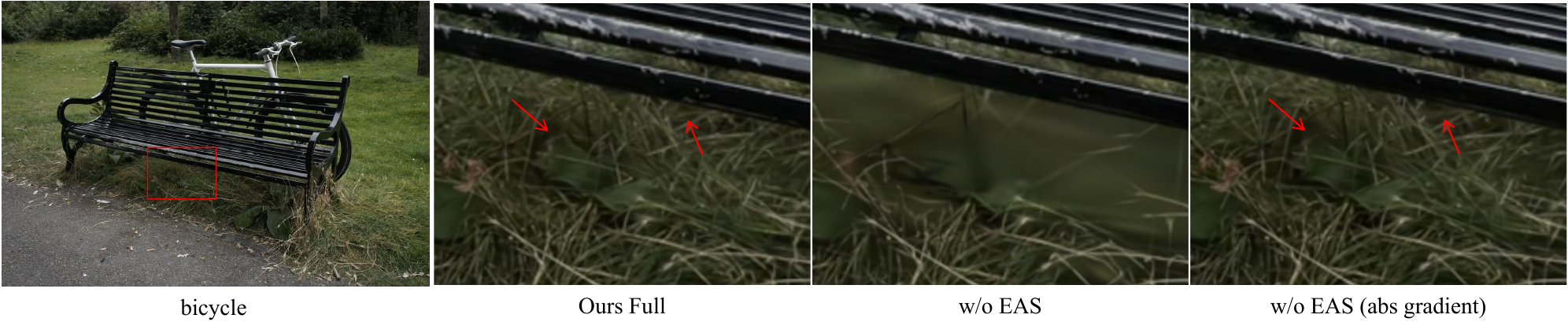}
    \caption{Qualitative comparison for evaluating the effectiveness of EAS.}
    \label{fig:test_EAS}
\end{figure*}

\subsection{Datasets and metrics}
\label{subsec:datasets}
We evaluated our method on real-world scenes from the Mip-NeRF 360 \cite{barron2022mip}, Tanks and Temples \cite{knapitsch2017tanks}, and Deep Blending \cite{hedman2018deep} datasets.
As with 3DGS, we selected all nine scenes from the Mip-NeRF 360 dataset, including five outdoor scenes and four indoor scenes.
For the Tanks and Temples dataset, we chose the train and truck scenes, and for the Deep Blending dataset, we selected the drjohnson and playroom scenes.
In each experiment, every 8th image was used as the validation set.
We report peak signal-to-noise ratio (PSNR), structural similarity (SSIM), and perceptual metric (LPIPS) from \cite{zhang2018unreasonable} as quality evaluation metrics.
All three scores were calculated using the methods provided by 3DGS.

\subsection{Implementation}
We built our code upon the open-source repository of TamingGS.
According to the scene size, we divided the 13 scenes into three categories, with each category having the same Gaussian budget.
The budget for bicycle, garden, and stump is 3M.
The budget for flowers, treehill, drjohnson, and truck is 1.5M.
The remaining scenes have a budget of 1M.

We reproduced 3DGS, AbsGS, PixelGS, MiniSplatting-D (MiniGS), TamingGS, 3DGS-MCMC, SteepGS, and Perceptual-GS as baseline methods, using all default parameter settings from their respective original papers.
All results represent the best performance obtained over three runs. All experiments were conducted on a single 4090D GPU.

\subsection{Qualitative Analysis}

The results of the comprehensive qualitative comparison are shown in Figure~\ref{fig:qualitative_comparison}, the results for the remaining 10 scenes can be found in Appendix A.3.

In the garden scene, our method significantly outperforms other works in terms of wall brick texture details and overall image cleanliness.
TamingGS achieves a relatively clean result but fails to recover detailed textures.
AbsGS is able to reconstruct most textures but suffers from poor image cleanliness.
3DGS-MCMC and SteepGS perform the worst: the former produces an overall blurry reconstruction, while the latter results in an unacceptably messy output.
Notably, only our method successfully recovers the white bricks on the right side of the wall.

In the drjohnson scene, our method delivers the highest quality in capturing fine details such as ceiling bulbs and sensors, while also maintaining the cleanest visual output.
MiniGS performs well in detail reconstruction; however, it uses several times more Gaussian budget than ours and shows inferior performance in image cleanliness.
PerceptualGS ranks second overall but struggles with large straight-line regions.
Other methods perform poorly, with 3DGS-MCMC completely missing many detailed areas.

In the train scene, only our method fully reconstructs the objects on the hillside.
All other methods exhibit noticeable blurred regions.
TamingGS ranks second in this case but still lags significantly behind our method.
Although SteepGS uses the least Gaussian budget, its reconstruction quality is the worst.

Overall, our method leads over other approaches in terms of image cleanliness, detail recovery, and budget efficiency.

\subsection{Quantitative Analysis}
The quantitative comparison results are shown in Table~\ref{tab:quantitative_results}.
Our method achieves the best results on all three datasets.
Among all compared methods, our approach uses the least Gaussian budget across all datasets, yet leads in both SSIM and PSNR metrics.
In terms of the LPIPS metric, our method is only slightly outperformed by MiniGS, which however employs nearly three times as many Gaussians as ours.
Although TamingGS and 3DGS-MCMC perform well in SSIM and PSNR, they both suffer from noticeable rendering quality issues: TamingGS struggles with certain blurred regions, while 3DGS-MCMC performs poorly in distant details.
Perceptual-GS is the most balanced among the baseline methods, but still lags significantly behind our approach.

Compared to works that achieve rendering quality better than 3DGS, we consistently achieve the best performance under the smallest budget in all scenes. 
Detailed per-scene metrics can be found in the Appendix A.1.

Due to differences in CUDA kernels used, it is not possible to directly compare training speed across different methods (e.g., the kernel used in TamingGS is significantly faster than that of 3DGS).
Under the same CUDA kernel, 3DGS-MCMC, PerceptualGS, and SteepGS require additional computational overhead, whereas other methods, including our approach, have computational costs proportional to the allocated Gaussian budget.
Compared to TamingGS, the fastest method in terms of training speed among the baselines (14.2 minutes), our approach further reduces the training time by half (6.7 minutes).

\subsection{Ablation Experiments}

\begin{table}[t]
	\centering
	\scalebox{0.85}{
		\begin{tabular}{l|ccccc}
			Method|Metric & $SSIM^\uparrow$ & $PSNR^\uparrow$ & $LPIPS^\downarrow$ & Time & FPS \\
			\hline
            Ours Full & 0.854 & 27.95 & 0.186 & 6.7 & 289 \\ 
            \hline
            w/o EAS & 0.837 & 27.77 & 0.220 & 6.4 & 346 \\ 
            w/o EAS(abs gradient) & 0.852 & 27.95 & 0.192 & 7.5 & 306 \\ 
            w/o LAS & 0.846 & 27.81 & 0.195 & 7.8 & 239 \\ 
            \hline
            w/o RAP & 0.851 & 27.79 & 0.187 & 7.8 & 267 \\ 
            w/o MU & 0.852 & 27.80 & 0.187 & 8.4 & 292 \\ 
            w/o GC & 0.853 & 27.94 & 0.188 & 7.8 & 301 \\ 
            w/o RAP\&MU\&GC & 0.848 & 27.67 & 0.193 & 9.1 & 286 
        \end{tabular}
	}
	\caption{Results of the ablation study on all datasets, where all configurations share the same budget.}
    \label{tab:ablation}
\end{table}

\begin{figure}[ht]
    \centering
    \includegraphics[width=0.45\textwidth]{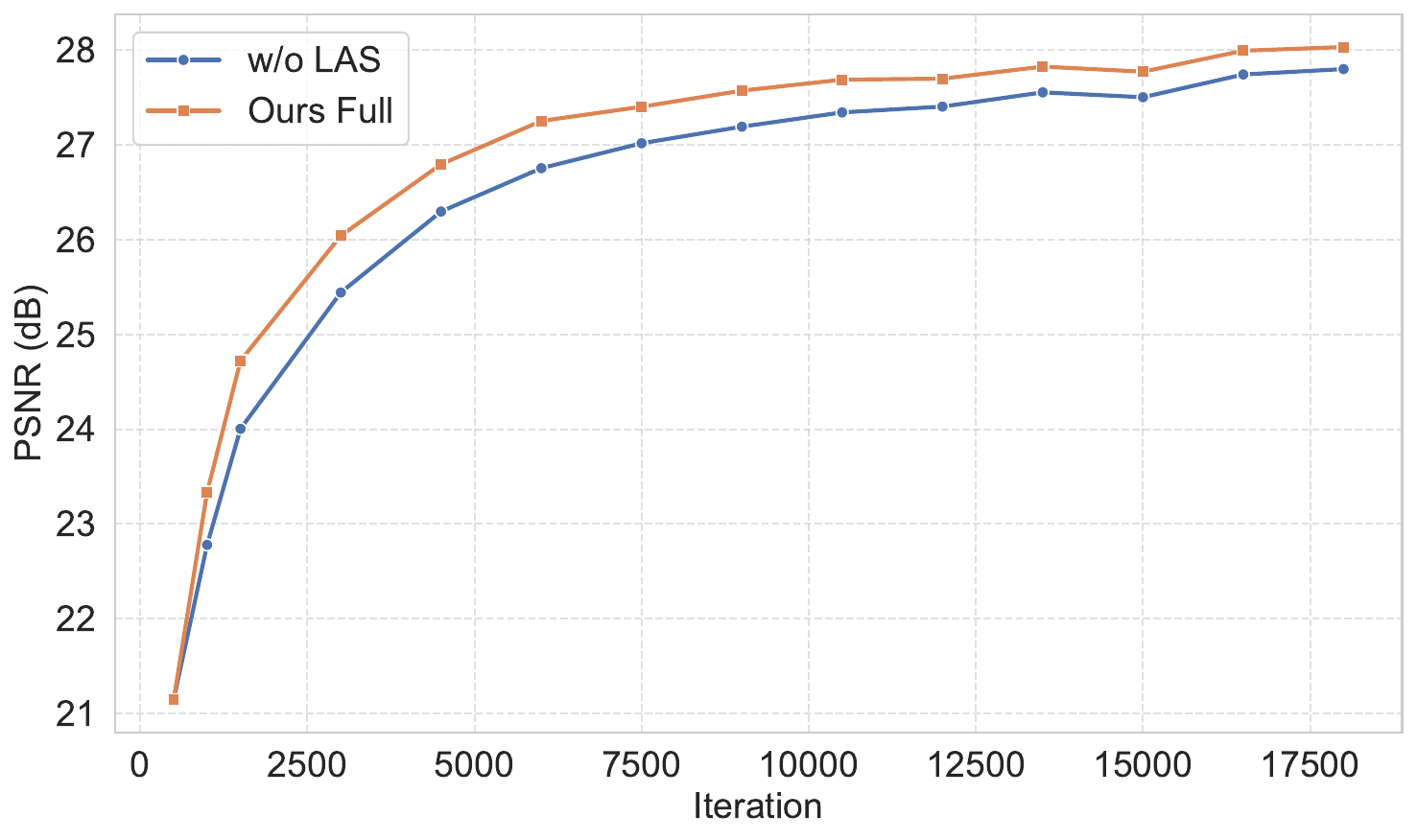}
    \caption{Evaluating the impact of LAS on optimization speed.}
    \label{fig:LAS_curve}
\end{figure}

\begin{figure}[ht]
    \centering
    \includegraphics[width=0.45\textwidth]{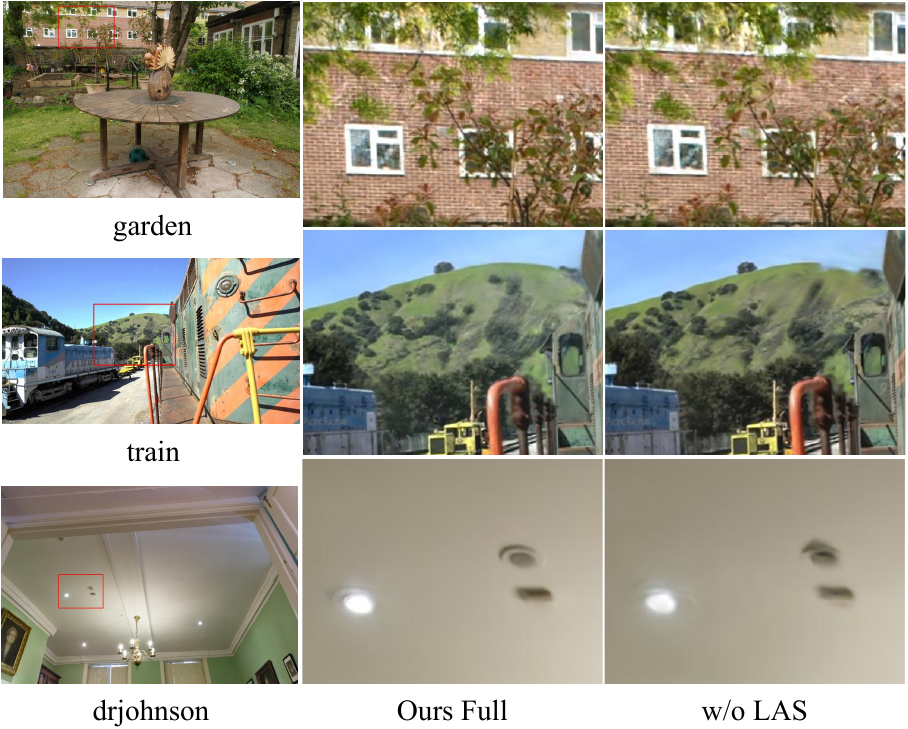}
    \caption{Qualitative comparison for evaluating the effectiveness of LAS. Only LAS enables the recovery of the white brick patch on the wall in the garden scene.}
    \label{fig:test_LAS}
\end{figure}

\begin{figure}[ht]
    \centering
    \includegraphics[width=0.45\textwidth]{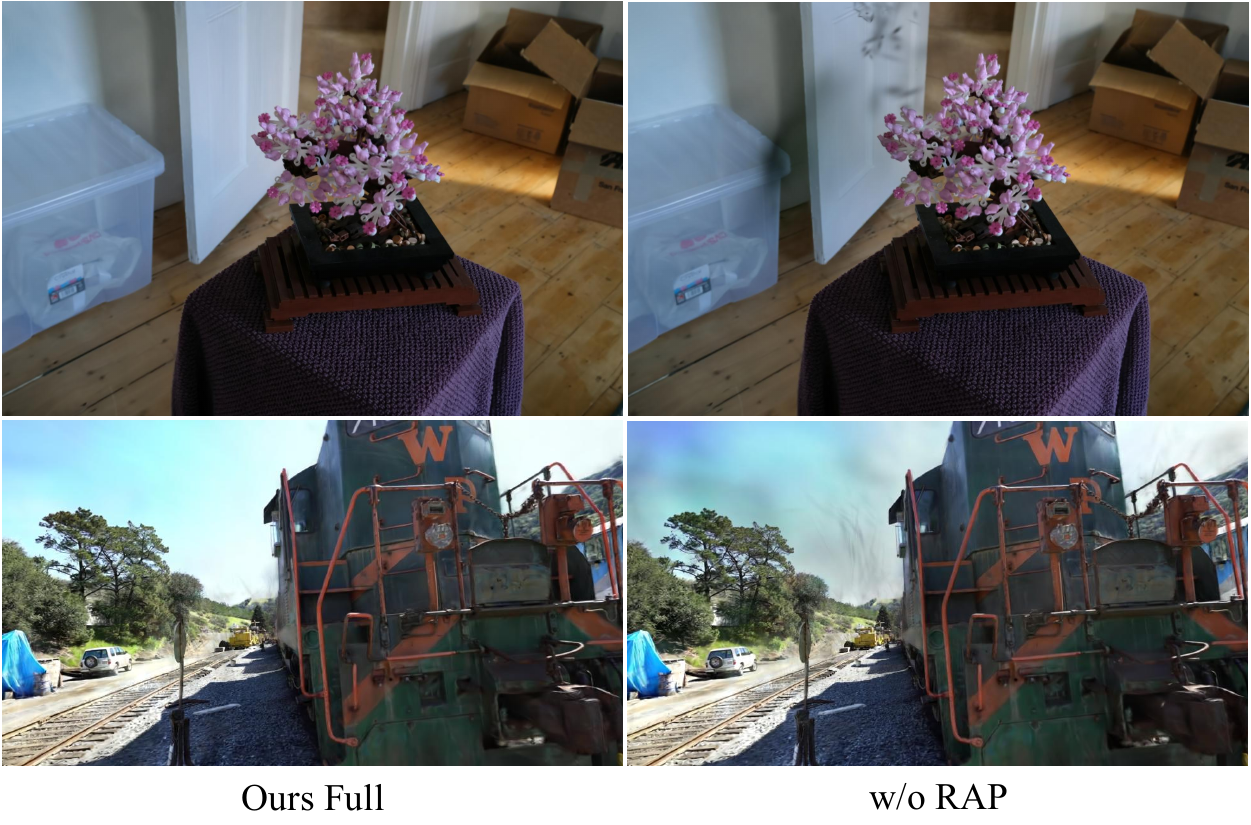}
    \caption{Qualitative comparison for evaluating the effectiveness of RAP.}
    \label{fig:test_RAP}
\end{figure}

\begin{figure}[ht]
    \centering
    \includegraphics[width=0.45\textwidth]{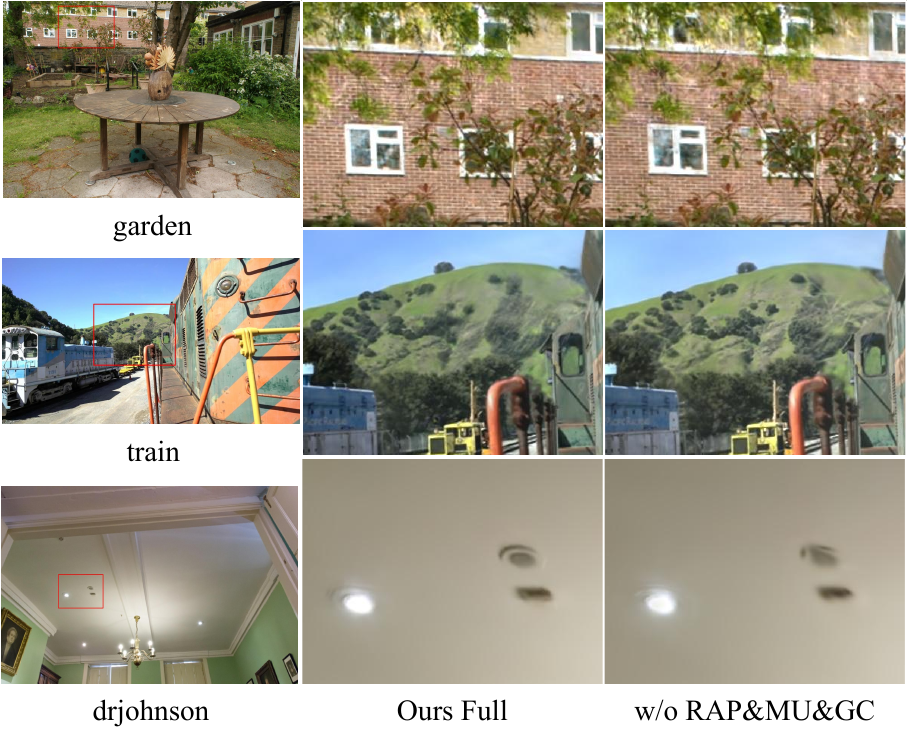}
    \caption{Qualitative comparison for evaluating the effectiveness of overfitting mitigation strategies.}
    \label{fig:test_RAP+MU+GC}
\end{figure}

The overall ablation study results are summarized in Table~\ref{tab:ablation}.
Since the ablation experiments do not change the CUDA kernel or the Gaussian budget, the training time and FPS metrics can be directly compared.

\textbf{Effect of EAS}: Refer to Rows 1, 2, and 3 in Table~\ref{tab:ablation}, EAS improves rendering quality, especially significantly in terms of the LPIPS metric.
The improvement in quality comes from two aspects: first, EAS performs well in identifying blurry regions; second, it enhances reconstruction quality at edge regions.
As shown in Figure~\ref{fig:test_EAS}, EAS helps eliminate reconstruction blur.
Compared to using only the absolute gradient, EAS achieves better reconstruction at leaf edges.
Although EAS increases rendering overhead, this is a natural trade-off for eliminating reconstruction blur.

\textbf{Effect of LAS}: Refer to Rows 1 and 4 in Table~\ref{tab:ablation}, LAS improves rendering quality while reducing rendering overhead.
LAS reduces rendering cost because, compared to clone and split, the resulting child Gaussians after LAS exhibit lower overlap ratios, thereby shortening the sequence of Gaussians involved in rendering each pixel.
The impact of Gaussian overlap on FPS can be found in Appendix A.8.
The improvement in rendering quality is attributed to the minimization of geometric discrepancies before and after splitting.
As illustrated in Figure~\ref{fig:LAS_curve}, LAS accelerates optimization, which aligns with our hypothesis.
As shown in Figure~\ref{fig:test_LAS}, LAS also improves detail reconstruction.

\textbf{Mitigating Overfitting}: Referring to the lower part of Table~\ref{tab:ablation}, we observe that RAP, MU, and GC all moderately enhance generalization performance and reduce training cost, with RAP also lowering rendering overhead.

RAP mainly improves performance by eliminating certain overfitted Gaussians, some of which contribute to floating artifacts that severely degrade rendering quality.
As seen in Figure~\ref{fig:test_RAP}, RAP effectively reduces the occurrence probability of such artifacts.
The reduction in rendering overhead may result from the removal of overfitted Gaussians and their descendants, which otherwise elongate the rendering sequence.

MU accelerates training by reducing the total number of parameter updates, while GC speeds up training by lowering the peak time of Gaussian count.

Comparing the first and last rows of Table~\ref{tab:ablation}, it is evident that the combined effect of these three overfitting mitigation components leads to a significant overall improvement.
The qualitative comparison can be referred to in Figure~\ref{fig:test_RAP+MU+GC}.

\section{Conclusion}
We optimized the densification process of 3D Gaussian Splatting (3DGS) from three perspectives: when to densify, how to densify, and how to alleviate overfitting.
The proposed method significantly improves rendering quality without introducing additional computational overhead.
Compared to current state-of-the-art densification approaches, our method achieves better reconstruction quality with fewer Gaussians.
Additionally, our work does not rely on specific CUDA kernels and can be easily integrated into existing 3DGS-based methods, making a significant contribution to the broader adoption of 3DGS.

\newpage
\bibliography{aaai2026}

\begin{thebibliography}{27}
\providecommand{\natexlab}[1]{#1}

\bibitem[{Barron et~al.(2022)Barron, Mildenhall, Verbin, Srinivasan, and Hedman}]{barron2022mip}
Barron, J.~T.; Mildenhall, B.; Verbin, D.; Srinivasan, P.~P.; and Hedman, P. 2022.
\newblock Mip-nerf 360: Unbounded anti-aliased neural radiance fields.
\newblock In \emph{Proceedings of the IEEE/CVF conference on computer vision and pattern recognition}, 5470--5479.

\bibitem[{Chen et~al.(2024)Chen, Wang, Wang, and Liu}]{chen2024text}
Chen, Z.; Wang, F.; Wang, Y.; and Liu, H. 2024.
\newblock Text-to-3d using gaussian splatting.
\newblock In \emph{Proceedings of the IEEE/CVF Conference on Computer Vision and Pattern Recognition}, 21401--21412.

\bibitem[{Cheng et~al.(2024)Cheng, Long, Yang, Yao, Yin, Ma, Wang, and Chen}]{cheng2024gaussianpro}
Cheng, K.; Long, X.; Yang, K.; Yao, Y.; Yin, W.; Ma, Y.; Wang, W.; and Chen, X. 2024.
\newblock Gaussianpro: 3d gaussian splatting with progressive propagation.
\newblock In \emph{Forty-first International Conference on Machine Learning}.

\bibitem[{Fang and Wang(2024)}]{fang2024mini}
Fang, G.; and Wang, B. 2024.
\newblock Mini-splatting: Representing scenes with a constrained number of gaussians.
\newblock In \emph{European Conference on Computer Vision}, 165--181. Springer.

\bibitem[{Fridovich-Keil et~al.(2022)Fridovich-Keil, Yu, Tancik, Chen, Recht, and Kanazawa}]{fridovich2022plenoxels}
Fridovich-Keil, S.; Yu, A.; Tancik, M.; Chen, Q.; Recht, B.; and Kanazawa, A. 2022.
\newblock Plenoxels: Radiance fields without neural networks.
\newblock In \emph{Proceedings of the IEEE/CVF conference on computer vision and pattern recognition}, 5501--5510.

\bibitem[{Hedman et~al.(2018)Hedman, Philip, Price, Frahm, Drettakis, and Brostow}]{hedman2018deep}
Hedman, P.; Philip, J.; Price, T.; Frahm, J.-M.; Drettakis, G.; and Brostow, G. 2018.
\newblock Deep blending for free-viewpoint image-based rendering.
\newblock \emph{ACM Transactions on Graphics (ToG)}, 37(6): 1--15.

\bibitem[{Kerbl et~al.(2023)Kerbl, Kopanas, Leimk{\"u}hler, and Drettakis}]{kerbl20233d}
Kerbl, B.; Kopanas, G.; Leimk{\"u}hler, T.; and Drettakis, G. 2023.
\newblock 3D Gaussian Splatting for Real-Time Radiance Field Rendering.
\newblock \emph{ACM Trans. Graph.}, 42(4): 139--1.

\bibitem[{Kheradmand et~al.(2024)Kheradmand, Rebain, Sharma, Sun, Tseng, Isack, Kar, Tagliasacchi, and Yi}]{kheradmand20243d}
Kheradmand, S.; Rebain, D.; Sharma, G.; Sun, W.; Tseng, Y.-C.; Isack, H.; Kar, A.; Tagliasacchi, A.; and Yi, K.~M. 2024.
\newblock 3d gaussian splatting as markov chain monte carlo.
\newblock \emph{Advances in Neural Information Processing Systems}, 37: 80965--80986.

\bibitem[{Knapitsch et~al.(2017)Knapitsch, Park, Zhou, and Koltun}]{knapitsch2017tanks}
Knapitsch, A.; Park, J.; Zhou, Q.-Y.; and Koltun, V. 2017.
\newblock Tanks and temples: Benchmarking large-scale scene reconstruction.
\newblock \emph{ACM Transactions on Graphics (ToG)}, 36(4): 1--13.

\bibitem[{Lee et~al.(2024)Lee, Lee, Sun, Ali, and Park}]{lee2024deblurring}
Lee, B.; Lee, H.; Sun, X.; Ali, U.; and Park, E. 2024.
\newblock Deblurring 3d gaussian splatting.
\newblock In \emph{European Conference on Computer Vision}, 127--143. Springer.

\bibitem[{Lin et~al.(2024)Lin, Dai, Zhu, and Yao}]{lin2024gaussian}
Lin, Y.; Dai, Z.; Zhu, S.; and Yao, Y. 2024.
\newblock Gaussian-flow: 4d reconstruction with dynamic 3d gaussian particle.
\newblock In \emph{Proceedings of the IEEE/CVF Conference on Computer Vision and Pattern Recognition}, 21136--21145.

\bibitem[{Mallick et~al.(2024)Mallick, Goel, Kerbl, Steinberger, Carrasco, and De~La~Torre}]{mallick2024taming}
Mallick, S.~S.; Goel, R.; Kerbl, B.; Steinberger, M.; Carrasco, F.~V.; and De~La~Torre, F. 2024.
\newblock Taming 3dgs: High-quality radiance fields with limited resources.
\newblock In \emph{SIGGRAPH Asia 2024 Conference Papers}, 1--11.

\bibitem[{Matsuki et~al.(2024)Matsuki, Murai, Kelly, and Davison}]{matsuki2024gaussian}
Matsuki, H.; Murai, R.; Kelly, P.~H.; and Davison, A.~J. 2024.
\newblock Gaussian splatting slam.
\newblock In \emph{Proceedings of the IEEE/CVF Conference on Computer Vision and Pattern Recognition}, 18039--18048.

\bibitem[{Mildenhall et~al.(2021)Mildenhall, Srinivasan, Tancik, Barron, Ramamoorthi, and Ng}]{mildenhall2021nerf}
Mildenhall, B.; Srinivasan, P.~P.; Tancik, M.; Barron, J.~T.; Ramamoorthi, R.; and Ng, R. 2021.
\newblock Nerf: Representing scenes as neural radiance fields for view synthesis.
\newblock \emph{Communications of the ACM}, 65(1): 99--106.

\bibitem[{M{\"u}ller et~al.(2022)M{\"u}ller, Evans, Schied, and Keller}]{muller2022instant}
M{\"u}ller, T.; Evans, A.; Schied, C.; and Keller, A. 2022.
\newblock Instant neural graphics primitives with a multiresolution hash encoding.
\newblock \emph{ACM transactions on graphics (TOG)}, 41(4): 1--15.

\bibitem[{Rota~Bul{\`o}, Porzi, and Kontschieder(2024)}]{rota2024revising}
Rota~Bul{\`o}, S.; Porzi, L.; and Kontschieder, P. 2024.
\newblock Revising densification in gaussian splatting.
\newblock In \emph{European Conference on Computer Vision}, 347--362. Springer.

\bibitem[{Schonberger and Frahm(2016)}]{schonberger2016structure}
Schonberger, J.~L.; and Frahm, J.-M. 2016.
\newblock Structure-from-motion revisited.
\newblock In \emph{Proceedings of the IEEE conference on computer vision and pattern recognition}, 4104--4113.

\bibitem[{Shao et~al.(2024)Shao, Wang, Li, Wang, Lin, Zhang, Fan, and Wang}]{shao2024splattingavatar}
Shao, Z.; Wang, Z.; Li, Z.; Wang, D.; Lin, X.; Zhang, Y.; Fan, M.; and Wang, Z. 2024.
\newblock Splattingavatar: Realistic real-time human avatars with mesh-embedded gaussian splatting.
\newblock In \emph{Proceedings of the IEEE/CVF Conference on Computer Vision and Pattern Recognition}, 1606--1616.

\bibitem[{Wang et~al.(2025)Wang, Wang, Wang, Mohan, Fan, Wu, Cai, Yeh, Wang, Liu et~al.}]{wang2025steepest}
Wang, P.; Wang, Y.; Wang, D.; Mohan, S.; Fan, Z.; Wu, L.; Cai, R.; Yeh, Y.-Y.; Wang, Z.; Liu, Q.; et~al. 2025.
\newblock Steepest Descent Density Control for Compact 3D Gaussian Splatting.
\newblock In \emph{Proceedings of the Computer Vision and Pattern Recognition Conference}, 26663--26672.

\bibitem[{Wu et~al.(2024)Wu, Yi, Fang, Xie, Zhang, Wei, Liu, Tian, and Wang}]{wu20244d}
Wu, G.; Yi, T.; Fang, J.; Xie, L.; Zhang, X.; Wei, W.; Liu, W.; Tian, Q.; and Wang, X. 2024.
\newblock 4d gaussian splatting for real-time dynamic scene rendering.
\newblock In \emph{Proceedings of the IEEE/CVF Conference on Computer Vision and Pattern Recognition}, 20310--20320.

\bibitem[{Yang et~al.(2024)Yang, Gao, Sun, Huang, Lyu, Zhou, Jiao, Qi, and Jin}]{yang2024spec}
Yang, Z.; Gao, X.; Sun, Y.-T.; Huang, Y.; Lyu, X.; Zhou, W.; Jiao, S.; Qi, X.; and Jin, X. 2024.
\newblock Spec-gaussian: Anisotropic view-dependent appearance for 3d gaussian splatting.
\newblock \emph{Advances in Neural Information Processing Systems}, 37: 61192--61216.

\bibitem[{Ye et~al.(2024)Ye, Li, Liu, Qiao, and Dou}]{ye2024absgs}
Ye, Z.; Li, W.; Liu, S.; Qiao, P.; and Dou, Y. 2024.
\newblock Absgs: Recovering fine details in 3d gaussian splatting.
\newblock In \emph{Proceedings of the 32nd ACM International Conference on Multimedia}, 1053--1061.

\bibitem[{Yu et~al.(2024)Yu, Chen, Huang, Sattler, and Geiger}]{yu2024mip}
Yu, Z.; Chen, A.; Huang, B.; Sattler, T.; and Geiger, A. 2024.
\newblock Mip-splatting: Alias-free 3d gaussian splatting.
\newblock In \emph{Proceedings of the IEEE/CVF Conference on Computer Vision and Pattern Recognition}, 19447--19456.

\bibitem[{Zhang et~al.(2018)Zhang, Isola, Efros, Shechtman, and Wang}]{zhang2018unreasonable}
Zhang, R.; Isola, P.; Efros, A.~A.; Shechtman, E.; and Wang, O. 2018.
\newblock The unreasonable effectiveness of deep features as a perceptual metric.
\newblock In \emph{Proceedings of the IEEE conference on computer vision and pattern recognition}, 586--595.

\bibitem[{Zhang et~al.(2024)Zhang, Hu, Lao, He, and Zhao}]{zhang2024pixel}
Zhang, Z.; Hu, W.; Lao, Y.; He, T.; and Zhao, H. 2024.
\newblock Pixel-gs: Density control with pixel-aware gradient for 3d gaussian splatting.
\newblock In \emph{European Conference on Computer Vision}, 326--342. Springer.

\bibitem[{Zhou and Ni(2025)}]{zhou2025perceptual}
Zhou, H.; and Ni, Z. 2025.
\newblock Perceptual-GS: Scene-adaptive Perceptual Densification for Gaussian Splatting.
\newblock \emph{arXiv preprint arXiv:2506.12400}.

\bibitem[{Zhou et~al.(2024)Zhou, Lin, Shan, Wang, Sun, and Yang}]{zhou2024drivinggaussian}
Zhou, X.; Lin, Z.; Shan, X.; Wang, Y.; Sun, D.; and Yang, M.-H. 2024.
\newblock Drivinggaussian: Composite gaussian splatting for surrounding dynamic autonomous driving scenes.
\newblock In \emph{Proceedings of the IEEE/CVF Conference on Computer Vision and Pattern Recognition}, 21634--21643.

\end{thebibliography}

\newpage
\appendix
\onecolumn
\section{Appendix}
\setlength{\tabcolsep}{6pt}

\subsection{Quantitative Comparison per Scene}

\begin{table}[ht]
	\centering
    \scalebox{0.82}{
	\begin{tabular}{l|c|c|c|c|c|c|c|c|c}
		Scene|Methods & 3DGS & AbsGS & PixelGS & MiniGS & TamGS & MCMC & SteepGS & PercepGS & Ours \\
        \hline
        bicycle & 0.76561 & 0.78243 & 0.77857 & 0.79877 & 0.77766 & 0.79896 & 0.73151 & 0.79283 & 0.80369  \\
        flowers & 0.60522 & 0.62010 & 0.63628 & 0.64182 & 0.61695 & 0.64659 & 0.54756 & 0.63873 & 0.63902  \\
        garden & 0.86670 & 0.86803 & 0.87039 & 0.87853 & 0.87378 & 0.87676 & 0.85584 & 0.87014 & 0.88139  \\
        stump & 0.77057 & 0.78059 & 0.78454 & 0.80582 & 0.77425 & 0.81188 & 0.73186 & 0.79787 & 0.80930  \\
        treehill & 0.63369 & 0.62100 & 0.63534 & 0.64207 & 0.64464 & 0.65881 & 0.61224 & 0.64299 & 0.65686  \\
        bonsai & 0.94217 & 0.94456 & 0.94657 & 0.94805 & 0.94532 & 0.94784 & 0.93827 & 0.94786 & 0.94911  \\
        counter & 0.90860 & 0.91116 & 0.91465 & 0.91044 & 0.91327 & 0.91709 & 0.90180 & 0.91520 & 0.91909  \\
        kitchen & 0.92771 & 0.93075 & 0.93117 & 0.93344 & 0.93180 & 0.93339 & 0.92294 & 0.93130 & 0.93617  \\
        room & 0.91915 & 0.92565 & 0.92242 & 0.92808 & 0.92460 & 0.92954 & 0.91558 & 0.92826 & 0.93194  \\
        \hline
        playroom & 0.90743 & 0.90871 & 0.90571 & 0.90753 & 0.90612 & 0.91606 & 0.90630 & 0.90715 & 0.91587  \\
        drjohnson & 0.90096 & 0.90194 & 0.88791 & 0.90505 & 0.90801 & 0.90809 & 0.90279 & 0.90454 & 0.91055  \\
        \hline
        train & 0.81321 & 0.82877 & 0.82760 & 0.82125 & 0.82614 & 0.83955 & 0.80120 & 0.82613 & 0.84640  \\
        truck & 0.88184 & 0.88549 & 0.88693 & 0.88939 & 0.89293 & 0.89926 & 0.87559 & 0.88798 & 0.89837  \\
        \end{tabular}}
	\caption{The SSIM scores for all works in each scene.}
\end{table}

\begin{table}[ht]
	\centering
    \scalebox{0.82}{
	\begin{tabular}{l|c|c|c|c|c|c|c|c|c}
		Scene|Methods & 3DGS & AbsGS & PixelGS & MiniGS & TamGS & MCMC & SteepGS & PercepGS & Ours \\
        \hline
        bicycle & 25.213 & 25.372 & 25.279 & 25.581 & 25.504 & 25.681 & 24.798 & 25.527 & 25.866   \\
        flowers & 21.539 & 21.368 & 21.580 & 21.526 & 21.871 & 22.010 & 20.713 & 21.494 & 21.768   \\
        garden & 27.361 & 27.408 & 27.493 & 27.693 & 27.898 & 27.811 & 27.125 & 27.628 & 28.098   \\
        stump & 26.539 & 26.726 & 26.843 & 27.140 & 26.632 & 27.384 & 25.832 & 27.030 & 27.211 \\
        treehill & 22.495 & 22.094 & 22.296 & 22.234 & 23.024 & 22.944 & 22.204 & 22.408 & 22.839 \\
        bonsai & 32.242 & 32.145 & 32.547 & 32.163 & 32.889 & 32.646 & 31.806 & 32.607 & 32.966 \\
        counter & 29.016 & 29.096 & 29.181 & 28.560 & 29.486 & 29.348 & 28.743 & 29.288 & 29.798 \\
        kitchen & 31.474 & 31.852 & 31.752 & 31.704 & 32.131 & 32.040 & 30.919 & 31.909 & 32.554 \\
        room & 31.446 & 31.634 & 31.588 & 31.528 & 32.199 & 32.188 & 31.330 & 32.040 & 32.652 \\
        \hline
        playroom & 30.019 & 30.051 & 29.876 & 30.447 & 30.186 & 30.449 & 30.099 & 30.200 & 30.627 \\
        drjohnson & 29.119 & 28.933 & 28.084 & 29.404 & 29.670 & 29.112 & 29.385 & 29.551 & 29.763 \\
        \hline
        train & 21.958 & 22.126 & 22.145 & 21.320 & 22.780 & 22.463 & 21.678 & 22.180 & 22.641 \\
        truck & 25.414 & 25.535 & 25.543 & 25.409 & 26.062 & 26.338 & 25.169 & 25.572 & 26.542 \\
        \end{tabular}}
	\caption{The PSNR scores for all works in each scene.}
\end{table}

\begin{table}[ht]
	\centering
    \scalebox{0.82}{
	\begin{tabular}{l|c|c|c|c|c|c|c|c|c}
		Scene|Methods & 3DGS & AbsGS & PixelGS & MiniGS & TamGS & MCMC & SteepGS & PercepGS & Ours \\
        \hline
        bicycle & 0.20921 & 0.18275 & 0.17956 & 0.15755 & 0.19210 & 0.16815 & 0.26239 & 0.17383 & 0.16510 \\
        flowers & 0.33536 & 0.28512 & 0.26170 & 0.25465 & 0.33052 & 0.28063 & 0.39758 & 0.26712 & 0.28529 \\
        garden & 0.10651 & 0.10663 & 0.09873 & 0.09011 & 0.09846 & 0.09575 & 0.12678 & 0.10273 & 0.09222 \\
        stump & 0.21659 & 0.20442 & 0.18780 & 0.16817 & 0.20476 & 0.17079 & 0.27371 & 0.18501 & 0.17718 \\
        treehill & 0.32461 & 0.29297 & 0.27505 & 0.26107 & 0.31206 & 0.27054 & 0.37381 & 0.28248 & 0.28717 \\
        bonsai & 0.20380 & 0.19269 & 0.19116 & 0.17372 & 0.19937 & 0.18995 & 0.21073 & 0.18098 & 0.18503 \\
        counter & 0.19967 & 0.19324 & 0.18275 & 0.17333 & 0.19408 & 0.18346 & 0.21453 & 0.17634 & 0.17865 \\
        kitchen & 0.12570 & 0.12130 & 0.11883 & 0.11396 & 0.12117 & 0.12040 & 0.13437 & 0.11704 & 0.11389 \\
        room & 0.21834 & 0.20404 & 0.20970 & 0.18750 & 0.20872 & 0.19769 & 0.22849 & 0.19437 & 0.19277 \\
        \hline
        playroom & 0.24304 & 0.24314 & 0.24037 & 0.20347 & 0.23718 & 0.23554 & 0.25294 & 0.23122 & 0.22560 \\
        drjohnson & 0.24406 & 0.24378 & 0.25469 & 0.21834 & 0.23513 & 0.23779 & 0.24825 & 0.23090 & 0.22588 \\
        \hline
        train & 0.20741 & 0.18948 & 0.17783 & 0.17908 & 0.20080 & 0.18502 & 0.22766 & 0.18306 & 0.16947 \\
        truck & 0.14644 & 0.13846 & 0.12024 & 0.10031 & 0.12577 & 0.11215 & 0.15890 & 0.11761 & 0.12091 \\
        \end{tabular}}
	\caption{The LPIPS scores for all works in each scene.}
\end{table}

\subsection{Implementation Details}
We built our code upon the TamingGS repository. In addition to the proposed method, we made the following modifications:
\begin{itemize}
  
      \item Regarding hyperparameters, we adjusted position lr init to 0.00004 and position lr final to 0.000002. The gradient threshold used in EAS is set to 0.0003 and the opacity reset threshold is set 0.05.
      \item We removed the pruning of large Gaussians, as our method can correctly identify these Gaussians and optimize their covered regions through splitting.
      
\end{itemize}

\subsection{More Qualitative Comparisons}

\begin{figure}[ht]
    \centering
    \includegraphics[width=0.93\textwidth]{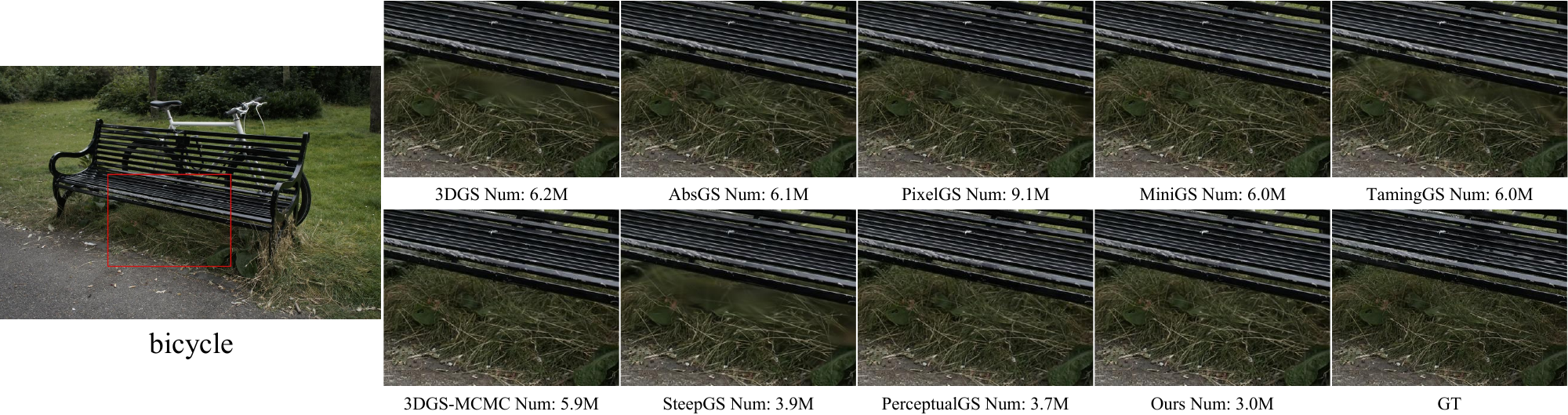} \\
    \includegraphics[width=0.93\textwidth]{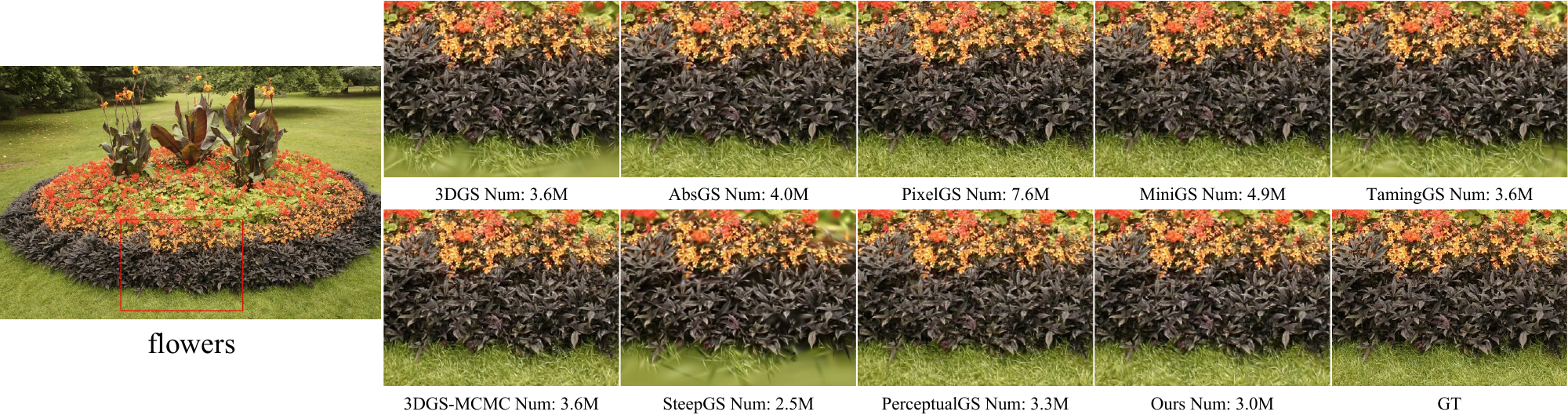} \\
    \includegraphics[width=0.93\textwidth]{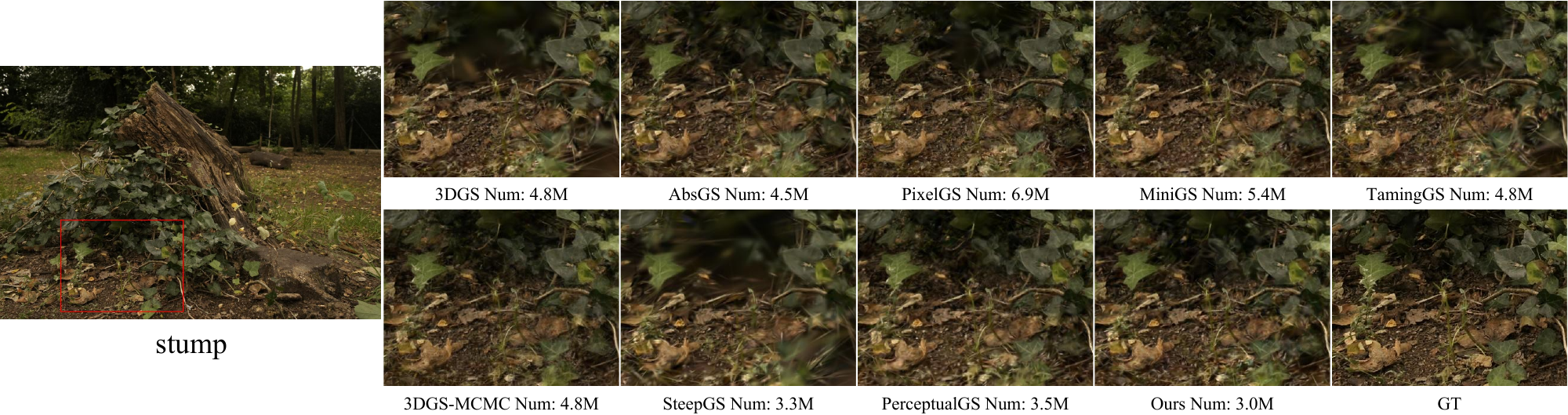} \\
    \includegraphics[width=0.93\textwidth]{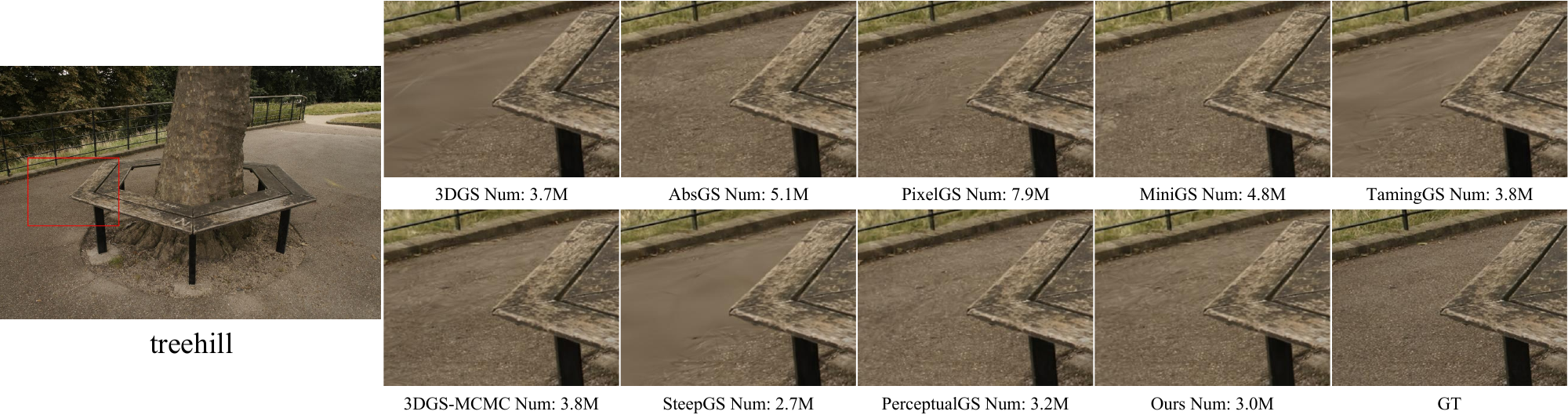} \\
    \includegraphics[width=0.93\textwidth]{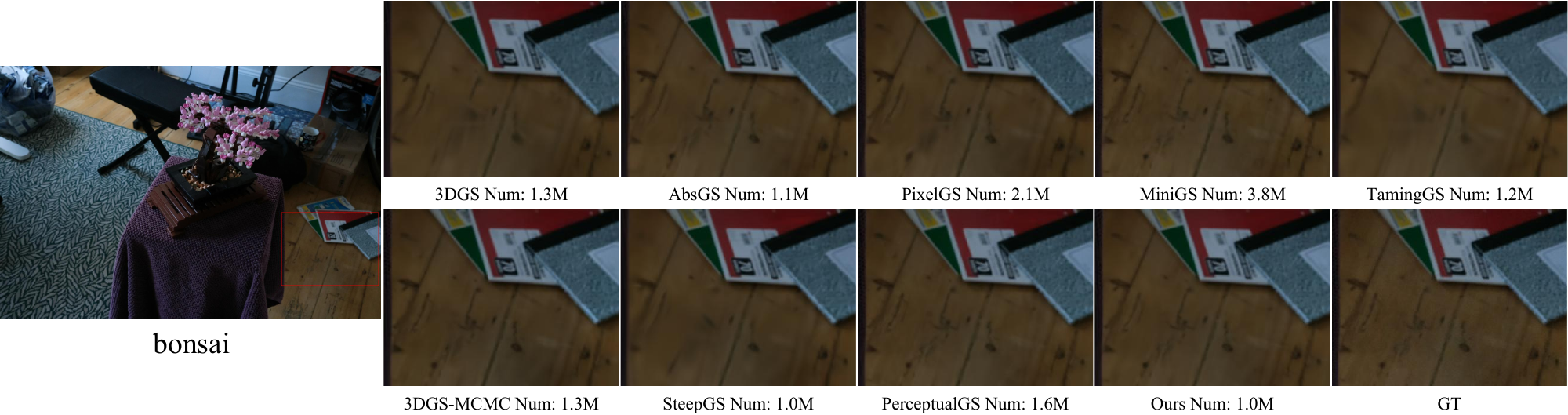} \\
    \caption{Qualitative comparison results among scenes bicycle, flowers, stump, treehill, bonsai.}
\end{figure}

\begin{figure}[ht]
    \centering
    \includegraphics[width=0.93\textwidth]{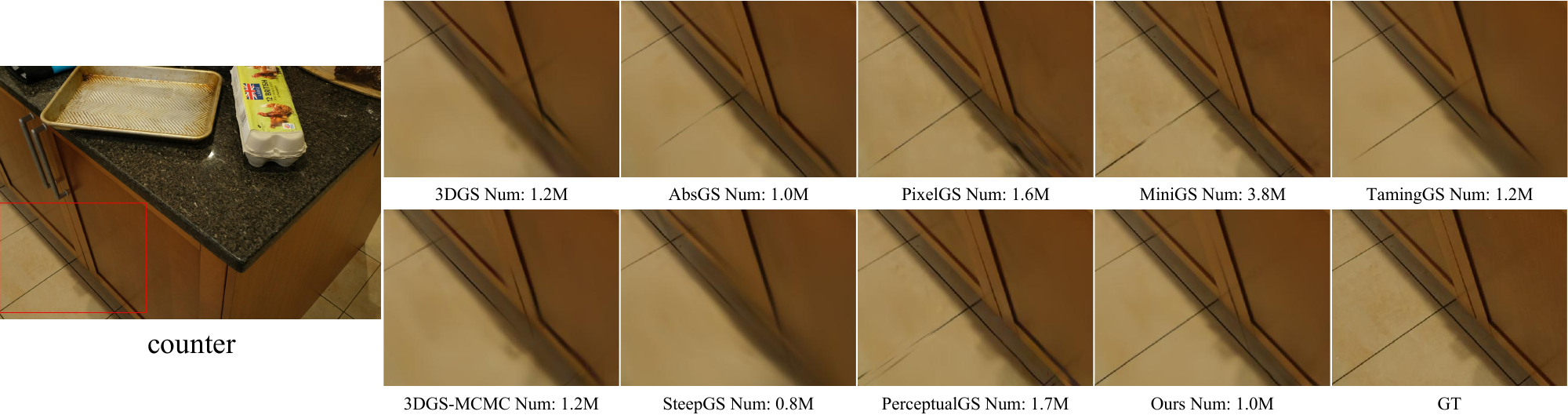} \\
    \includegraphics[width=0.93\textwidth]{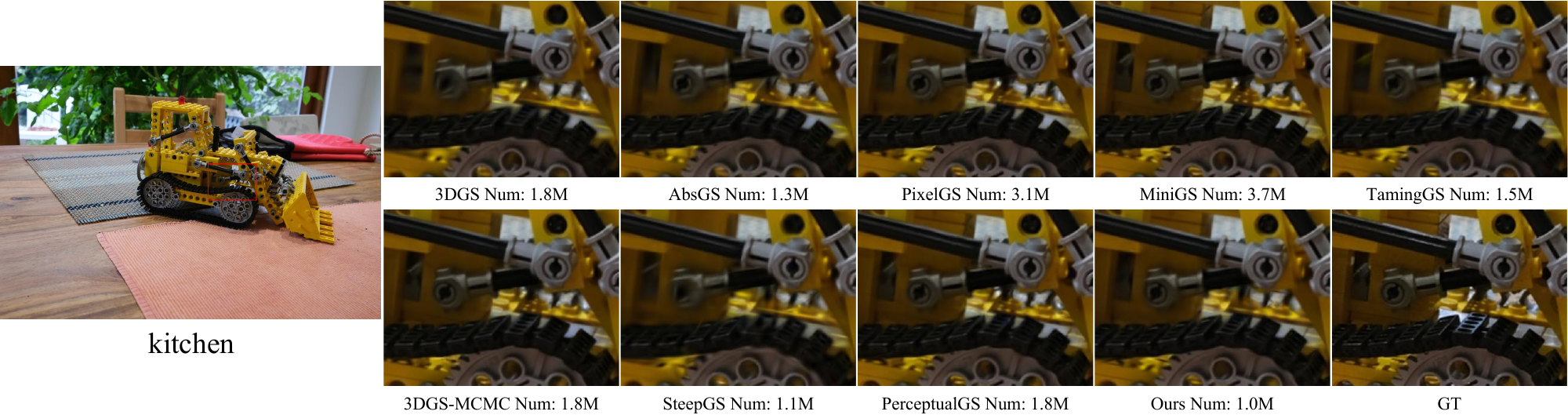} \\
    \includegraphics[width=0.93\textwidth]{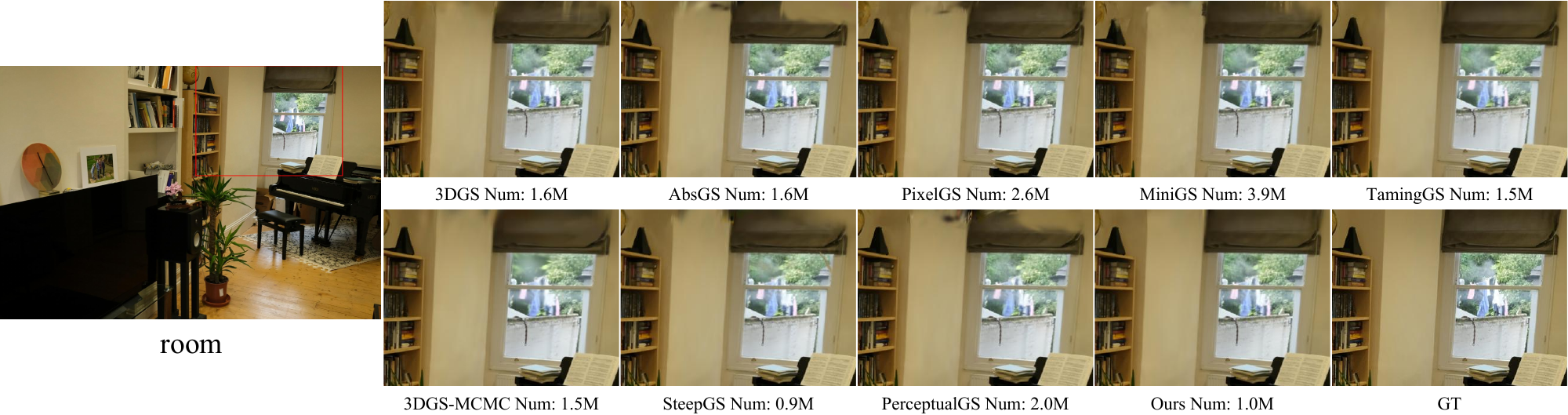} \\
    \includegraphics[width=0.93\textwidth]{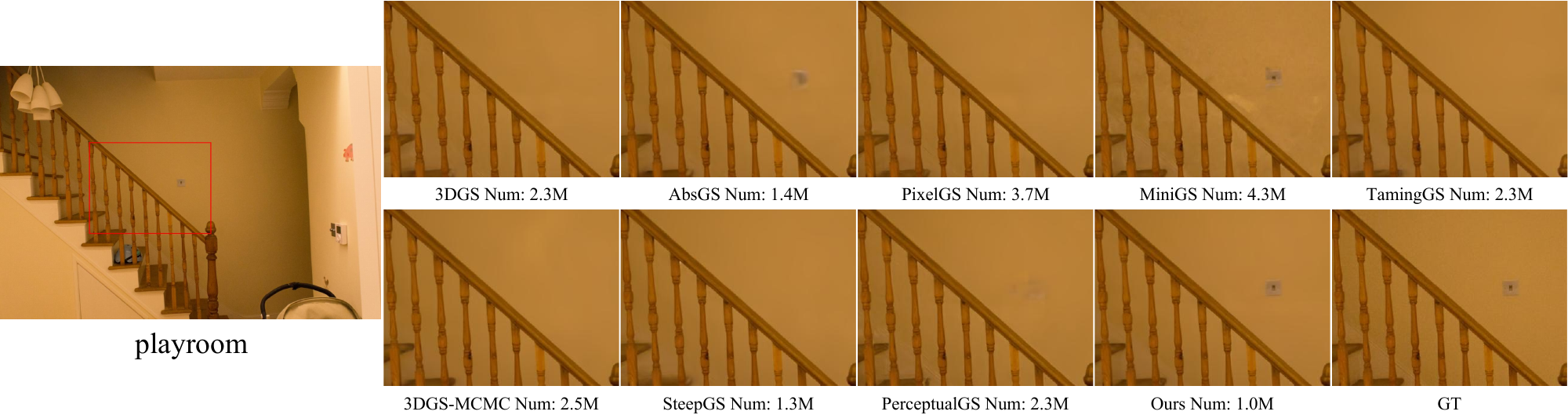} \\
    \includegraphics[width=0.93\textwidth]{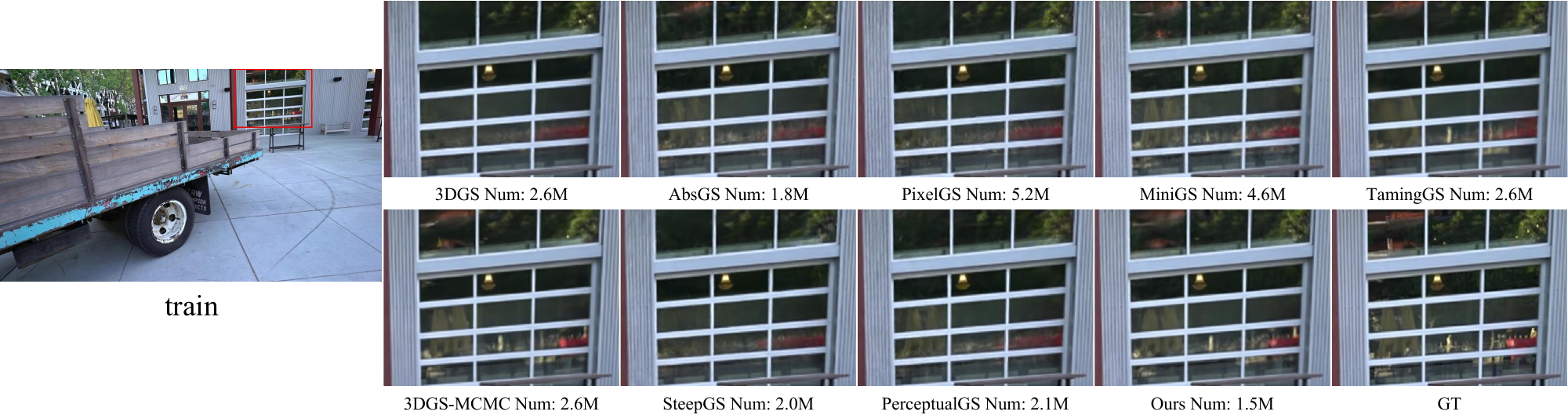} \\
    \caption{Qualitative comparison results among scenes counter, kitchen, room, playroom, truck.}
\end{figure}

\clearpage

\subsection{Proof Regarding the Optimal Rs}

\begin{figure}[ht]
    \centering
    \includegraphics[width=0.7\textwidth]{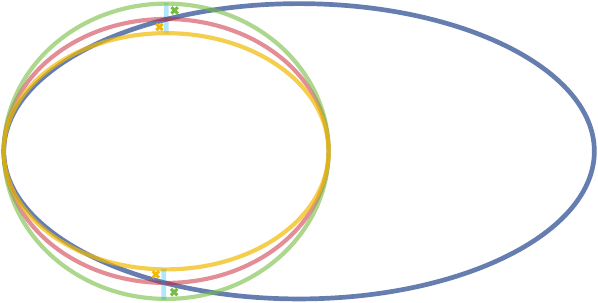}
    \caption{Schematic illustration of the split operation when simplified to a 2D ellipse.}
    \label{fig:proof}
\end{figure}

We can first simplify the shape change of a 3D Gaussian after splitting into a 2D ellipse form (since the scaling ratios along the two shorter axes are identical, excluding the longest axis).
Through intuitive geometric comparison(see Figure~\ref{fig:proof}), it is straightforward to prove that when the condition in the formula(red lines):
\begin{equation}
R_s = R_0 \cdot \sqrt{1 - \frac{d^2}{L_0^2}},
\end{equation}
the absolute difference between the overlapping and non-overlapping areas of the child and parent Gaussians is minimized. 

In this case, if we increase the length of $R_s$ (green lines), the increment in the overlapping area is smaller than the increment in the non-overlapping area.
The regions marked with green X in the figure represent the additional non-overlapping area compared to the overlapping area.
Conversely, if we decrease the length of $R_s$ (yellow lines), the reduction in the overlapping area exceeds the reduction in the non-overlapping area.
The regions marked with yellow "X" indicate the extra area lost in the overlap relative to the non-overlap.
The actual introduced area difference is twice the size of these marked regions.

\begin{figure}[ht]
    \centering
    \includegraphics[width=0.6\textwidth]{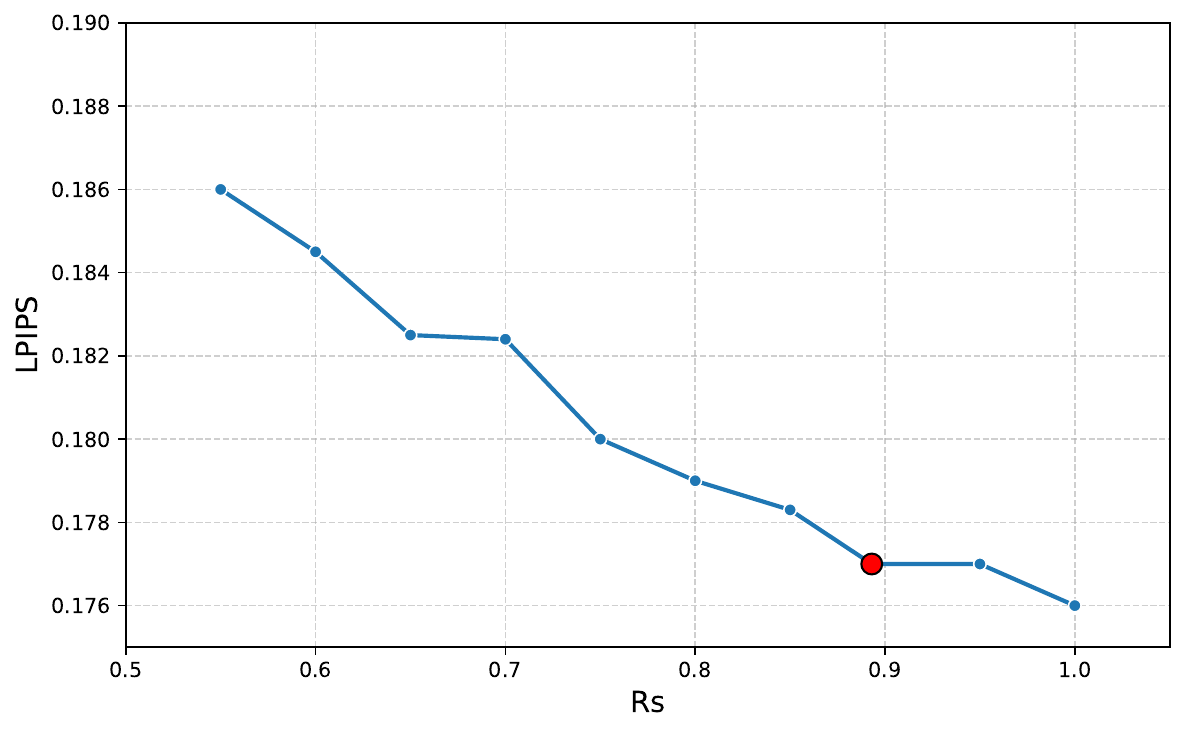}
    \caption{The trend of LPIPS with respect to Rs in the stump scene. The red dot indicates the value we chose in practice.}
    \label{fig:rs}
\end{figure}

In practice, the geometric error introduced by splitting also depends on factors such as color, opacity, and other Gaussians covering the same region.
Therefore, while the geometric area-optimal solution may not be absolutely optimal, it is very close to the global optimum.
Figure~\ref{fig:rs} provides a comparison of results on the "stump" scene using different values of $R_s$.

\subsection{Quantitative Comparison Under the Same Budget}
TamingGS and 3DGS-MCMC are the only two methods among the compared approaches that allow manually setting a budget for the number of Gaussians.
Here, we also provide quantitative results for these two methods under the same Gaussian budget as our method.

\begin{table}[ht]
	\centering
	\scalebox{0.95}{
		\begin{tabular}{l|cccc}
			Dataset & \multicolumn{4}{c}{Average} \\
            \hline
			Method|Metric
			& $SSIM^\uparrow$   & $PSNR^\uparrow$    & $LPIPS^\downarrow$  & $Num^\downarrow$\\
            \hline
            TamingGS(SIGGRAPHAsia24) & 0.83777 & 27.673 & 0.21659 & 1615312\\ 
            3DGS-MCMC(NeurIPS24) & 0.84969 & 27.706 & 0.20075 & 1615385\\ 
            Ours & 0.85367 & 27.948 & 0.18609 & 1615385 
        \end{tabular}
	}
	\caption{Quantitative comparison across all scenes.}
\end{table}

\subsection{Ablation Study of MU}

\begin{figure}[!ht]
    \centering
    \includegraphics[width=0.8\textwidth]{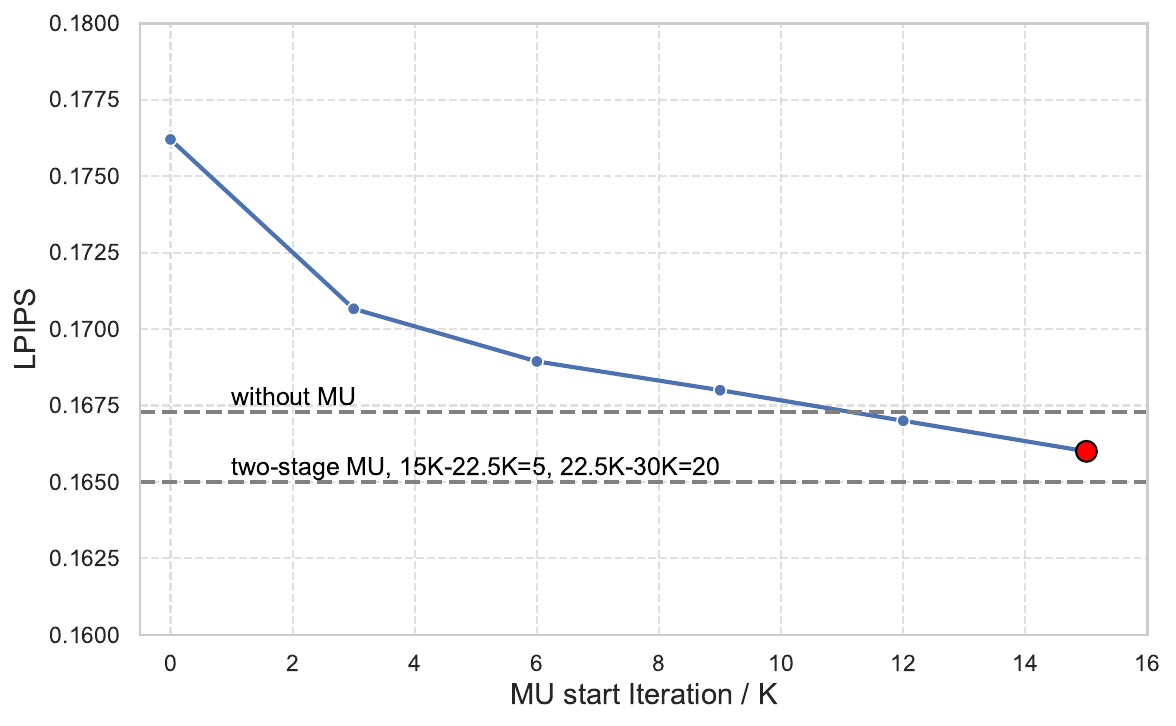}
    \caption{The impact of different MU starting iterations on LPIPS, test scene is bicycle. The update interval for parameters is uniformly set to 5. The red dot indicates the value we chose in practice.}
    \label{fig:mu}
\end{figure}

Figure~\ref{fig:mu} shows the impact of enabling MU at different times.
According to the figure, enabling MU from the beginning significantly degrades rendering quality.
The negative impact of MU on rendering quality gradually diminishes as the densification process progresses.
Enabling MU at the end of the densification process  can improve rendering quality.
Our two-stage MU approach performs slightly better than the single-stage MU.

\subsection{Ablation Study of Opacity Reduction Rate}

\begin{figure}[!ht]
    \centering
    \includegraphics[width=0.8\textwidth]{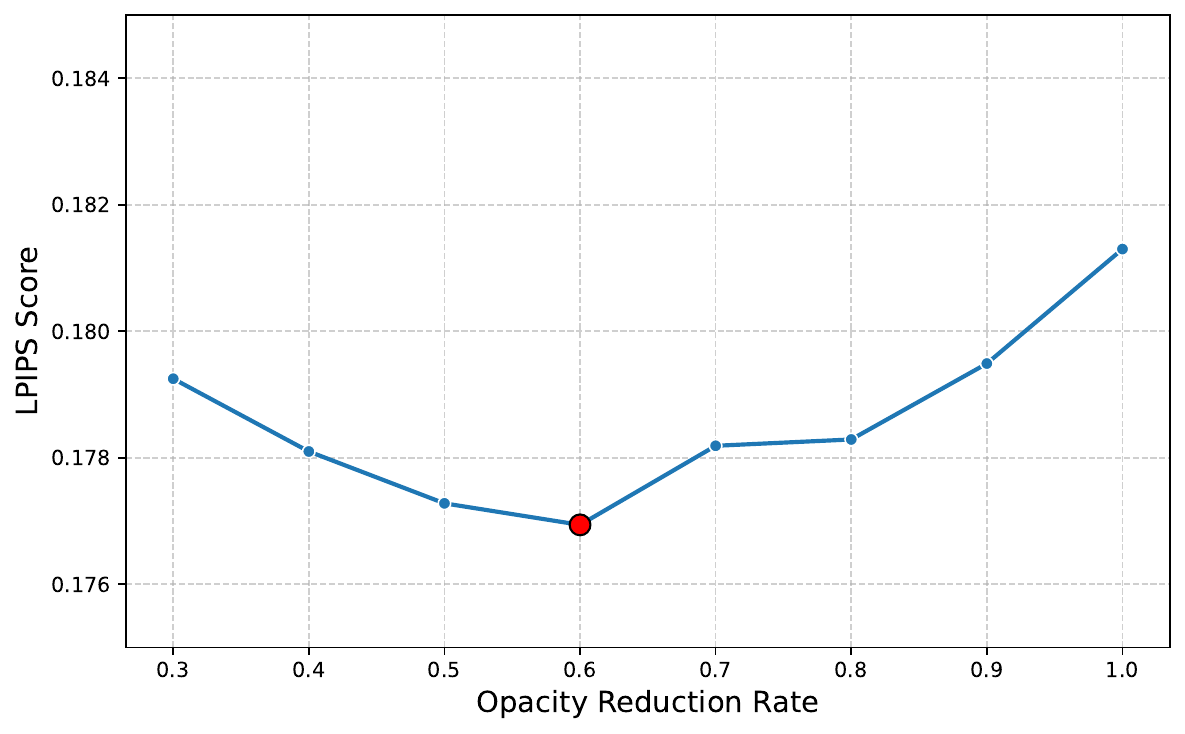}
    \caption{The trend of LPIPS with respect to opacity reduction rate in the stump scene. The red dot indicates the value we chose in practice.}
    \label{fig:opacity_reduction}
\end{figure}

Figure~\ref{fig:opacity_reduction} shows the impact of different Opacity Reduction Rates on rendering quality.
As can be seen from the figure, reducing opacity after splitting improves rendering quality, regardless of the reduction magnitude.
When the Rate varies between 0.4 and 0.8, the change in rendering quality is relatively small.
We choose 0.6 as an empirically determined optimal value.

\subsection{Ablation Study of Split Distance}

\begin{figure}[!ht]
    \centering
    \includegraphics[width=0.8\textwidth]{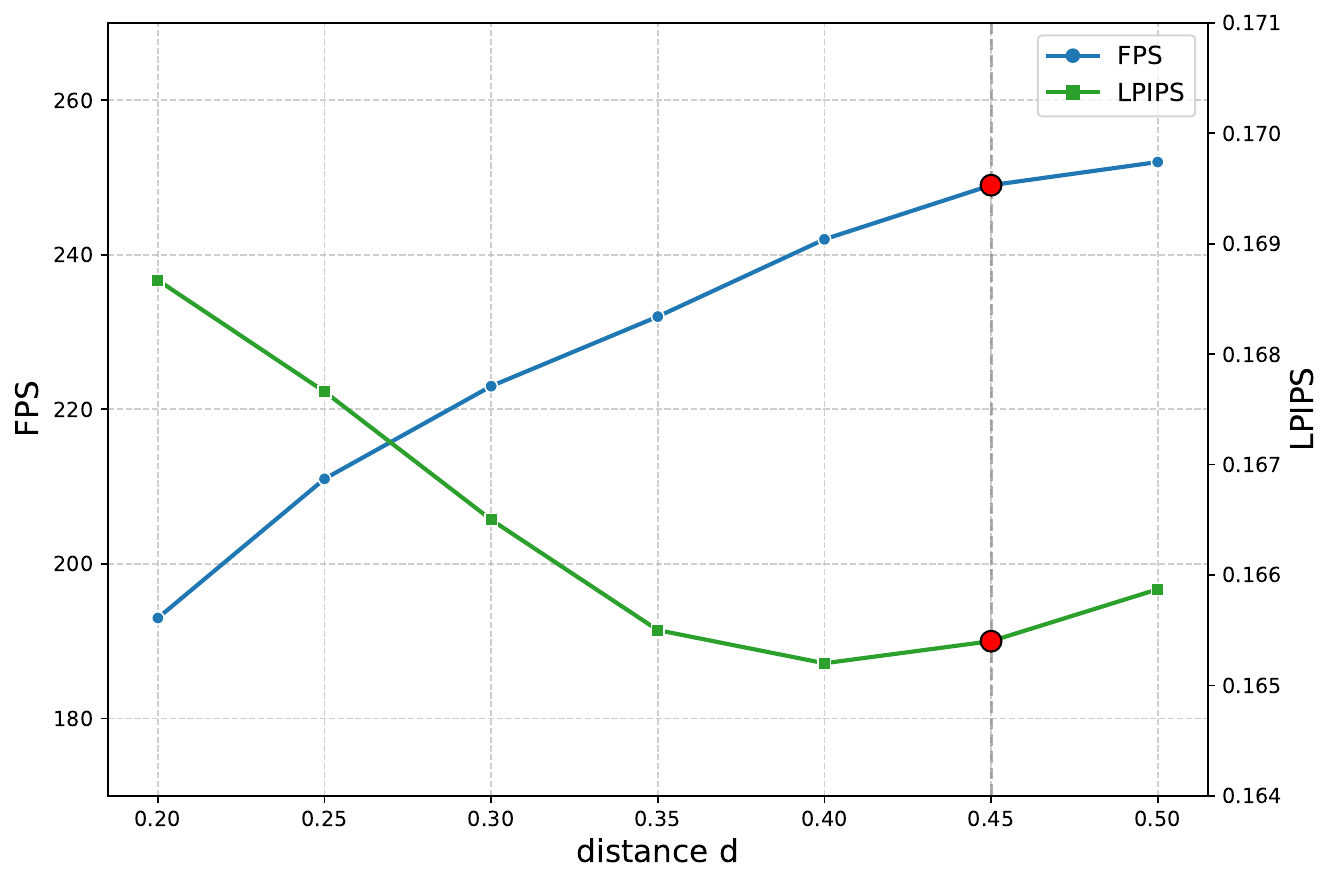}
    \caption{Testing the impact of the distance d between the sub-Gaussian and the original Gaussian centers on quality and rendering frame rate, test scene is bicycle.}
    \label{fig:test_d}
\end{figure}

Figure~\ref{fig:test_d} shows the impact of the distance (d) between child Gaussians and the original Gaussian center after splitting, on both rendering quality and FPS.
As d gradually decreases from 0.5, the geometric discrepancy caused by splitting slightly reduces, but the overlapping area among Gaussians increases significantly.
This overlap increases the pixel rendering queue length, leading to higher rendering costs.
Moreover, we observe that as d decreases, the negative effect of Gaussian overlap on rendering quality gradually outweighs the positive effect from reduced geometric discrepancy.
The value of 0.45 we choose is an empirically determined trade-off that balances FPS and rendering quality.

\subsection{Ablation Study of Growth Control}

\begin{figure}[!ht]
    \centering
    \includegraphics[width=0.8\textwidth]{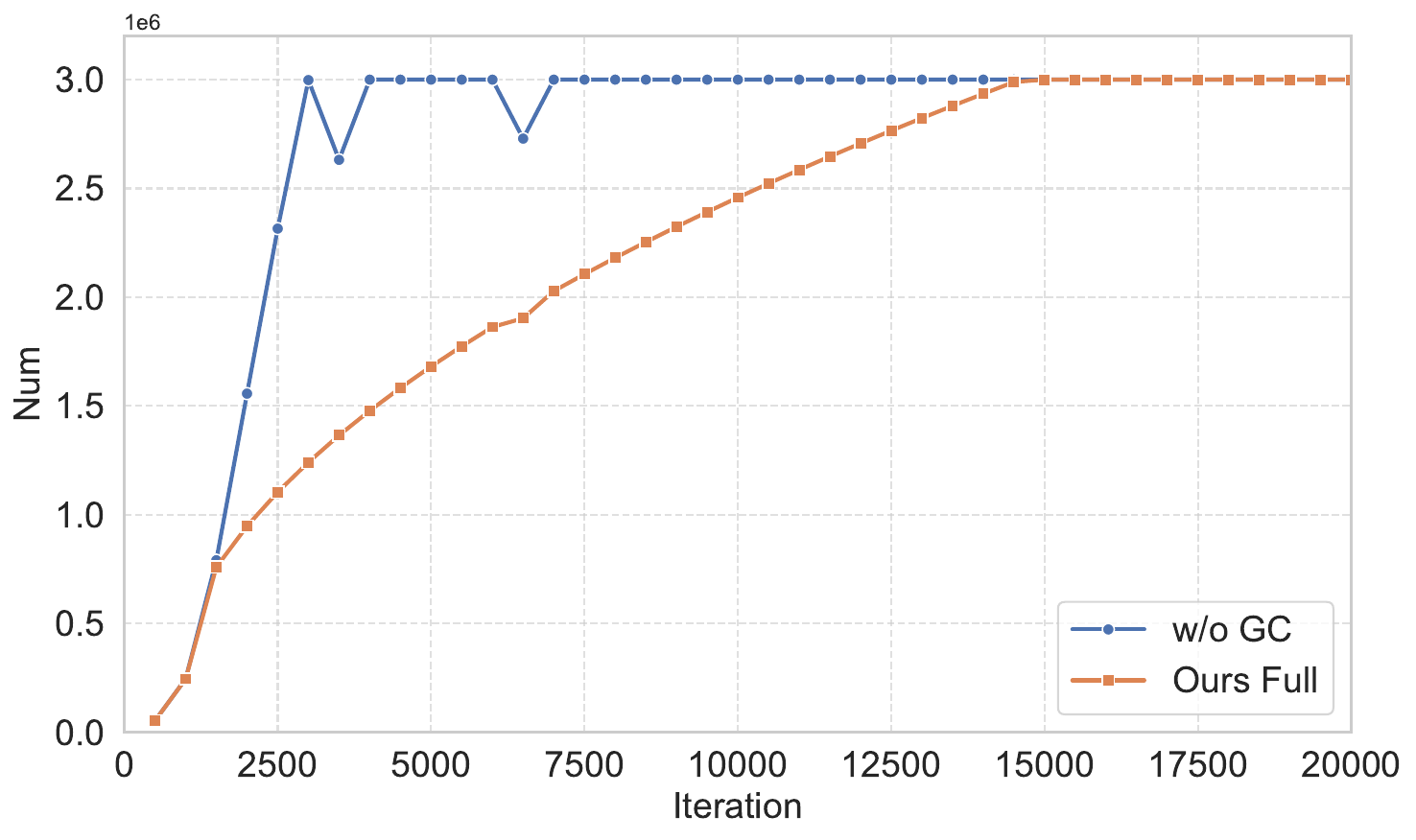}
    \caption{Growth curves of Gaussians with and without GC, test scene is bicycle.}
    \label{fig:gc}
\end{figure}

Figure~\ref{fig:gc} shows the growth curves of Gaussians with and without GC.
Without GC, the number of Gaussians reaches its peak early in the densification process, which clearly increases rendering overhead.
In contrast, with GC applied, the growth of Gaussians follows a smoother trend, reaching its peak precisely at the end of densification.

\end{document}